\documentclass[sigconf]{acmart}

\copyrightyear{2024}
\acmYear{2024}
\setcopyright{acmlicensed}
\acmConference[CIKM '24] {Proceedings of the 33rd ACM International Conference on Information and Knowledge Management}{October 21--25, 2024}{Boise, ID, USA.}
\acmBooktitle{Proceedings of the 33rd ACM International Conference on Information and Knowledge Management (CIKM '24), October 21--25, 2024, Boise, ID, USA}
\acmISBN{979-8-4007-0436-9/24/10}
\acmDOI{10.1145/3627673.3679760}
\usepackage{booktabs} 
\usepackage{siunitx} 
\usepackage{multirow}
\usepackage{multicol}
\usepackage{graphicx}
\usepackage{subcaption}
\usepackage{booktabs}
\usepackage{enumitem}

\AtBeginDocument{%
  }

\begin{document}

\title{SINKT: A Structure-Aware Inductive Knowledge Tracing Model \\with Large Language Model}

\author{Lingyue Fu}
\affiliation{%
  \institution{Shanghai Jiao Tong University}
  \city{Shanghai}
  \country{China}
}
\email{fulingyue@sjtu.edu.cn}

\author{Hao Guan}
\affiliation{%
  \institution{Shanghai Jiao Tong University}
  \city{Shanghai}
  \country{China}
}
\email{Guanhao2022@sjtu.edu.cn}

\author{Kounianhua Du}
\affiliation{%
  \institution{Shanghai Jiao Tong University}
  \city{Shanghai}
  \country{China}
}
\email{kounianhuadu@sjtu.edu.cn}

\author{Jianghao Lin}
\affiliation{%
  \institution{Shanghai Jiao Tong University}
  \city{Shanghai}
  \country{China}
}
\email{chiangel@sjtu.edu.cn}

\author{Wei Xia}
\affiliation{
\institution{Huawei Noah's Ark Lab}
\city{Shenzhen}
\country{China}
}
\email{xiawei24@huawei.com}

\author{Weinan Zhang}
\affiliation{%
  \institution{Shanghai Jiao Tong University}
  \city{Shanghai}
  \country{China}
}
\email{wnzhang@sjtu.edu.cn}

\author{Ruiming Tang}
\affiliation{
\institution{Huawei Noah's Ark Lab}
\city{Shenzhen}
\country{China}
}
\email{tangruiming@huawei.com}

\author{Yasheng Wang}
\affiliation{
\institution{Huawei Noah's Ark Lab}
\city{Shenzhen}
\country{China}
}
\email{wangyasheng@huawei.com}

\author{Yong Yu}
\affiliation{%
  \institution{Shanghai Jiao Tong University}
  \city{Shanghai}
  \country{China}
}
\email{yyu@apex.sjtu.edu.cn}
\authornote{The corresponding author.}

\renewcommand{\shortauthors}{Fu et al.}

\newcommand{\dk}[1]{{\color{blue}[dk: #1]}}
\newcommand{\jh}[1]{{\color{red}[jh: #1]}}
\newcommand{\fly}[1]{{\color{green}[fly: #1]}}
\begin{abstract}
Knowledge Tracing (KT) aims to determine whether students will respond correctly to the next question, which is a crucial task in intelligent tutoring systems (ITS). In educational KT scenarios, transductive ID-based methods often face severe data sparsity and cold start problems, where interactions between individual students and questions are sparse, and new questions and concepts consistently arrive in the database. In addition, existing KT models only implicitly consider the correlation between concepts and questions, lacking direct modeling of the more complex relationships in the heterogeneous graph of concepts and questions.  In this paper, we propose a \underline{S}tructure-aware \underline{IN}ductive \underline{K}nowledge \underline{T}racing model with large language model (dubbed \textbf{SINKT}), which, for the first time, introduces large language models (LLMs) and realizes inductive knowledge tracing. Firstly, SINKT utilizes LLMs to introduce structural relationships between concepts and constructs a heterogeneous graph for concepts and questions. Secondly, by encoding concepts and questions with LLMs, SINKT incorporates semantic information to aid prediction. Finally, SINKT predicts the student's response to the target question by interacting with the student's knowledge state and the question representation. Experiments on four real-world datasets demonstrate that SINKT achieves state-of-the-art performance among 12 existing transductive KT models. Additionally, we explore the performance of SINKT on the inductive KT task and provide insights into various modules.
  
\end{abstract}





\maketitle
\section{Introduction}
Intelligent tutoring systems (ITS), such as Massive Open Online Courses (MOOCs)\footnote{https://www.mooc.org/} and Khan Academy\footnote{https://www.khanacademy.org/}, are garnering increasing attention from learners due to the extensive learning resources and real-time feedback mechanisms.
Knowledge Tracing (KT) stands out as a vital area of research within ITS, which aims to assess the current knowledge states of students by analyzing their learning histories and to predict their responses to upcoming questions. KT not only helps to identify deficiencies in student's learning content but also provides a foundational basis for the subsequent instructional strategies in ITS.

Existing KT models originate from traditional methods represented by BKT~\cite{bkt} and have evolved significantly with the advent of models based on deep neural networks.
Deep learning-based KT models learn ID embeddings for questions and concepts through learning history of students.  
DKT~\cite{dkt} and DHKT~\cite{dhkt} utilize Recurrent Neural Networks (RNNs)~\cite{rnn} to capture the sequential information of learning history. 
EERNNA~\cite{eernna}, SAKT~\cite{sakt}, and AKT~\cite{akt} employ attention mechanisms to gauge the importance of a student’s learning history in addressing the current question.
SKVMN~\cite{skvmn} and DKVMN~\cite{dkvmn} apply memory networks~\cite{weston2015memory} to exploit the relationships between concepts and questions.
Some models~\cite{iekt,cl4kt,lbkt,gkt,gikt} introduce additional information from the learning process to assist response prediction.
IEKT~\cite{iekt} selects learning histories of similar students to enhance the understanding of the current student’s knowledge state.
 GKT~\cite{gkt} and GIKT~\cite{gikt} incorporate the transition graph originated from learning history and use Graph Neural Networks (GNNs)~\cite{kipf2016semi} to extract representations from graph data. 
 CL4KT~\cite{cl4kt} designs positive and negative sample pairs and employs contrastive learning~\cite{hadsell2006dimensionality} to generate more accurate representations of questions.
LBKT~\cite{lbkt} introduces various learning behaviors, including speed, attempts, and hints, to identify the learning patterns of students. Despite the success of existing KT models, some important limitations still exist. 

\begin{figure}[t]
    \centering
    \includegraphics[width=0.65\linewidth]{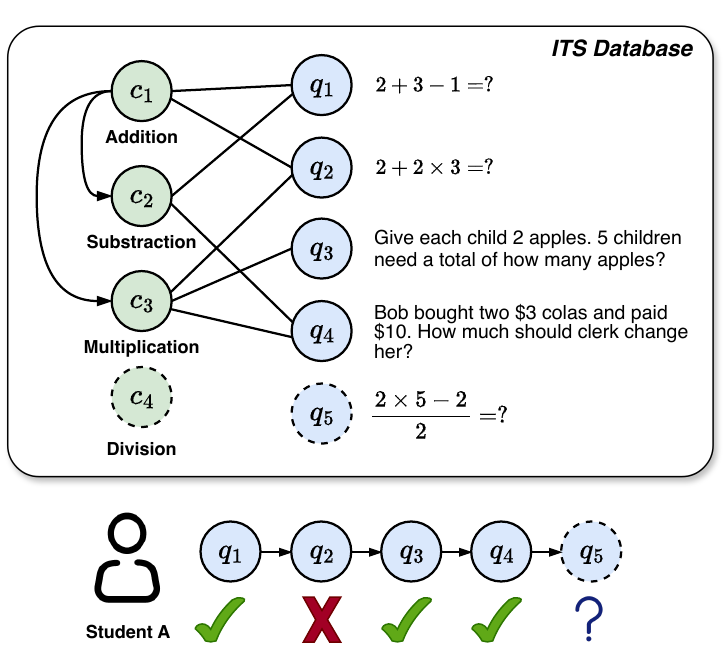}
    \caption{A demonstration of duplicated relations between concepts and questions in ITS. Concept $c_4$ and question $q_5$ are newly added to ITS, which do not appear in any learning history.}
    \label{fig:intro}
\end{figure}

Firstly, current ID-based transductive KT models cannot give accurate predictions to newly arrived questions and questions with sparse interactions.
\textit{Transductive KT models} learn ID embeddings of questions and concepts from learning history and predict responses to those questions that have appeared in the dataset. Existing KT models are all ID-based and are therefore considered transductive KT models. 
However, as illustrated in Figure \ref{fig:intro}, when a new concept $c_4$ and a new question $q_5$ are added to the database of ITS, transductive KT models lack interaction data to learn their representations, thus unable to predict the response of the student A.
Models that could predict those questions that haven't appear in the dataset are called \textit{inductive KT models}.
Meanwhile, interactions with individual students are limited, and thus question-level information is very sparse, which is difficult for ID-based transductive KT models to train. 
For example, the commonly used MovieLens dataset~\footnote{https://grouplens.org/datasets/movielens/} in recommendation systems contains 25,000,095 ratings and 1,093,360 tag applications across 62,423 movies. However, in educational datasets, such as ASSIST09\footnote{https://sites.google.com/site/assistmentsdata/home/2009-2010-assistment-data}, there are only 2,661 students with 165,455 interactions, while the total number of questions is 14,083.

Secondly, current methods fail to capture the semantic dependencies and the sufficient topological relations among questions and concepts. For example, as shown in Figure~\ref{fig:intro}, $q_1$ and $q_2$ are computational questions, while $q_3$ and $q_4$ are application questions. Compared to computational questions, application questions require students to possess additional analytical skills and have higher cognitive demands. Additionally,  structural information exists among concepts. For instance, addition $c_1$ should serve as the foundation for subtraction $c_2$ and multiplication $c_3$ during the learning process. If a student struggles with the concept ``addition'', he will likely have poor demand for concepts ``subtraction'' and ``multiplication''. Directly learning from learning sequences using ID embeddings makes it difficult to capture such type of information. In ITS, structural and semantic information are indispensable, as they can provide additional clues for knowledge tracing.

Based on these considerations, in this paper, we propose a novel \underline{S}tructure-Aware \underline{IN}ductive \underline{K}nowledge \underline{T}racing Model with Large Language Model, named \textbf{SINKT}. To the best of our knowledge, SINKT is the first work that can achieve predictions for newly ingested questions, \textit{i.e.,} realize inductive knowledge tracing. Our framework also integrates open-world semantic and structural information by LLMs for the first time, enabling it to dynamically update and expand its knowledge base with minimal manual intervention. 
Specifically, SINKT first employs LLMs to generate a concept-question heterogeneous graph that contains useful structural information in ITS.
Then, we use a Pretrained Language Model (PLM) to encode semantic information of questions and concepts instead of training ID embeddings.
Subsequently, we encode the concept-question graph by applying a carefully designed structural information encoder.
Finally, we design a student state encoder to capture sequential information and an interaction predictor to predict the student's response with his knowledge state and the target question description. 

To sum up, our contributions are summarized as follows:

\begin{itemize}[leftmargin=10pt]
    \item We propose a novel Structure-Aware Inductive Knowledge Tracing model with LLM (\textit{i.e.,} SINKT), and a full-automated pipeline for ITS to inductively trace mastery of new concepts and predict response to new questions, which is the first model to realize inductive knowledge tracing.
    \item We utilize LLMs for the first time to carefully inject open-world knowledge, including semantic and structural information of questions and concepts, into the KT task.
    \item Extensive experiments on four real-world datasets demonstrate that SINKT achieves state-of-the-art performance against 12 baselines, as well as superior capabilities to predict responses for unseen questions.
\end{itemize}

\section{RELATED WORK}

\subsection{Knowledge Tracing}
To investigate the learning patterns of students in ITS, numerous KT models have been proposed to encode knowledge and model interaction histories.
Traditional KT models such as Bayesian Knowledge Tracing~\cite{corbett1994knowledge} focus on tracing students' knowledge states by estimating general parameters. 
With the advent of deep learning, DKT~\cite{dkt} is the first to utilize RNN~\cite{rnn} and LSTM~\cite{hochreiter1997long} to model the sequential dynamics of student interactions.
DKT+~\cite{yeung2018addressing} and DHKT~\cite{dhkt} extend the implementation of DKT, taking into account the relationships between questions and concepts.
Inspired by the Transformer architecture~\cite{vaswani2017attention}, attention mechanisms~\cite{attention} have been incorporated into deep learning KT models for capturing the relationships among questions and their relevance to a student’s knowledge states~\cite{akt, sakt, eernna}.

Recently, attention has been paid to more nuanced behavioral data and advanced neural architectures. For instance, LBKT~\cite{lbkt} analyzes a range of learning behaviors, containing the speed of responses, the number of attempts, and the use of hints. SAKT~\cite{sakt} and AKT~\cite{akt} introduce attention mechanisms~\cite{attention} to focus on the most relevant aspects of a student's past interactions. Additionally, innovative approaches such as CMKT~\cite{cmkt}assess students' dynamic mastery of concepts. MF-DAKT~\cite{mfakt}, which incorporates multiple student-related factors with a dual attention mechanism, showcases the field's advancement towards more granular and complex models. 
Alongside these developments, graph-based KT models like GKT~\cite{gkt} and GIKT~\cite{gikt} employ GNN~\cite{kipf2016semi} to leverage the relational data among concepts. 

\subsection{LLM-Enhanced User Modeling}
There have been some works using LLMs to enhance the performance of user modeling~\cite{lin2023can,tan2023user}.
 Some work~\cite{lin2024rella, yang2023palr,liu2023once,Liu2023AFL} apply LLMs as profilers to involve the creation of prompts based on users' history, and input these prompts into LLMs to generate various aspects of user profiles. KAR~\cite{xi2023openworld} leverages LLMs to generate user and item profiles, encompassing user preference and real-world knowledge of items into click-through rate (CTR) prediction task.
GIRL~\cite{zheng2023generative} uses LLMs to generate suitable job descriptions base on the user's curriculum vitae to help recommendation model understand jobhunter's preferences.
In addition to profiling, LLMs serve as feature encoders, translating raw user data into rich, semantic embeddings that improve user modeling systems~\cite{10.1007/978-3-031-40292-0_13,li2023llm4jobs,xi2023openworld}.  GPT4SM~\cite{10.1007/978-3-031-40292-0_13} employs GPT to encode queries and candidate text for relevance prediction in recommendation systems, and LKPNR~\cite{hao2023lkpnr} uses open-source LLMs for better semantic representation of news articles. 
Moreover, LLMs function as knowledge augmenters by integrating external knowledge into user models~\cite{xi2023openworld,li2023gpt4rec,li2023prompt,Acharya2023LLMBG,fang2023chatgpt}, thus enhancing the precision of user-related predictions. \citet{mysore2023large} augment narrative-driven recommendations by generating author narrative queries with LLMs, and KAR~\cite{xi2023openworld} uses LLM-prompted outputs to augment recommender systems with factual item knowledge.

Existing LLM-enhanced user modeling methods are mostly applied in the domain of recommender systems (RS), which differ significantly from the KT task. In RS, users' interests typically remain stable over short periods, whereas in ITS, the knowledge states of students evolve dynamically as they engage with educational contents such as exercises. Additionally, there is strong structural information, such as interdependency and correlation among the concepts in ITS. Therefore, there is a need for specially designed models that can effectively track and adapt to the unique dynamics of students’ learning processes in ITS.

\begin{figure*}[t]
    \centering
    \includegraphics[width=0.93\linewidth]{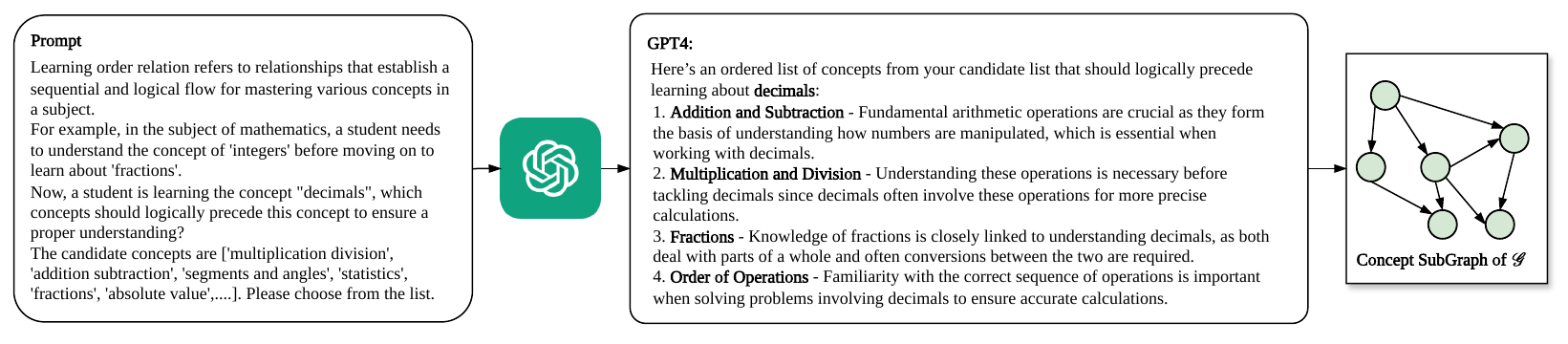}
    \caption{Prompt demonstration of concept relation graph generation. }
    \label{fig:prompt}
\end{figure*}

\section{Preliminaries}
\subsection{Problem Definition}
Consider a set of students $\mathcal{S}$, a set of questions $\mathcal{Q}$, and a set of concepts $\mathcal{C}$ within an ITS. The learning history of a student $s \in \mathcal{S}$ is recorded as $R_s = \{(q_1, r_1), (q_2, r_2), \dots, (q_T, r_T)\}$, where $q_t \in \mathcal{Q}$ represents the question answered by the student at time step $t$, and $r_t \in \{0,1\}$ indicates whether the student $s$ correctly answers the question $q_t$ (1 for correct and 0 for incorrect).
Given the learning history $R_s$ of a student and a new question $q_{T+1}$, the goal of the KT task is to predict the probability that the student correctly answers the new question, denoted as $p(r_{T+1} = 1 | R_s, q_{T+1})$.

Note that in a real ITS, teachers continually expand the question and concept set, meaning that both the question set $\mathcal Q$ and the concept set $\mathcal C$ are in a state of ongoing growth. 
 Previous KT models are trained and tested under the assumption $\mathcal Q_{train} = \mathcal{Q}_{test}$, which are only suitable for the transductive KT task. However, in the inductive KT task, $\mathcal Q_{train} \not= \mathcal Q_{test}.$ In this paper, our SINKT is capable of addressing the inductive task effectively while also delivering excellent performance in the transductive KT task.

\vspace{-11pt}
\subsection{Concept-Question Graph Generation}
To inject open-world knowledge into SINKT, we construct a heterogeneous concept-question graph $\mathcal{G}=(\mathcal{V}, \mathcal E, \mathcal{O_V}, \mathcal{R_E})$ to capture effective representations of concepts and questions, where $\mathcal{O_V}$ represent the set of vertex type, and $\mathcal{R_E}$ denotes the edge type (relation). 
The vertex set $\mathcal V$ includes the question set $\mathcal Q$ and the concept set $\mathcal{C}$, and The vertex type set $\mathcal{O_V}$ includes \textit{concept} and \textit{question} correspondingly.
Edge set $\mathcal E$ describes different relationships between concepts and questions. The edge type set $\mathcal{R_E}$ includes \textit{concept-question}, \textit{question-concept} and \textit{concept-concept}.
The graph $\mathcal{G}$ not only encapsulates the relationships between questions and concepts contained in the dataset but also incorporates open-world knowledge provided by LLMs.

\textit{concept-question} edges and \textit{question-concept} edges are derived from the dataset.  
Each question $q_i$ corresponds to one or more concepts $\mathcal C_{q_i} = \{c_1, c_2, \dots, c_{n_i}\}$, and each concept $c_i$ corresponds to many questions $\mathcal Q_{c_i} = \{q_1, q_2, \dots, q_{m_i}\}$, where $n_i$ and $m_i$ denotes the number of concepts related to the question $q_i$ and the number of questions related to the concept $c_i$, respectively. In the graph $\mathcal G$, the relation is described as $n_i$ \textit{question-concept} edges $\langle q_i, c_j\rangle (j = 1,2,\dots, n_i)$ and $m_i$ \textit{concept-question} edges $\langle c_i, q_j\rangle  (j =1,2,\dots, m_i)$.

\textit{concept-concept} edges within the graph $\mathcal G$ are generated by LLM.
As demonstrated in Figure \ref{fig:prompt}, for each concept $c_i$, we employ a specific prompt and utilize GPT-4~\cite{openai2023gpt4} to identify a list of related concepts. 
\cite{gpt4graph2023, modality2023} suggest that LLMs struggle with tasks directly involving graph structures, often requiring specialized training or input transformation to handle graph structures. Therefore, we opt to use a ``select from the list'' approach rather than asking GPT-4 to directly generate the graph.
After GPT-4 generates the response, we use the regular expression extraction to get $p_i$ concepts related to $c_i$, and add the directed edges $\langle c_j, c_i\rangle (j=1,2,\dots, p_i)$ into the graph $\mathcal{G}$.

\section{Methodology}

\begin{figure*}[t]
    \centering
    \begin{subfigure}{.29\textwidth}
    \centering
    \includegraphics[width=\linewidth]{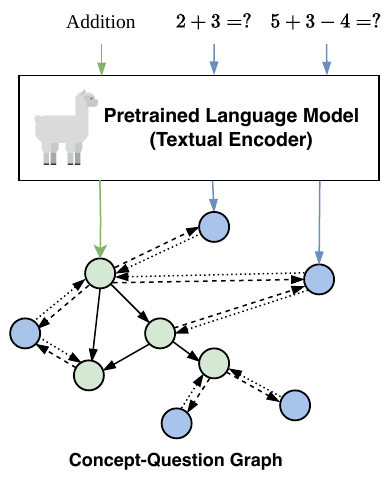}
    \caption{Textual Information Encoder (TIEnc).}
    \end{subfigure}
    \begin{subfigure}{.35\textwidth}
        \centering
        \includegraphics[width=\linewidth]{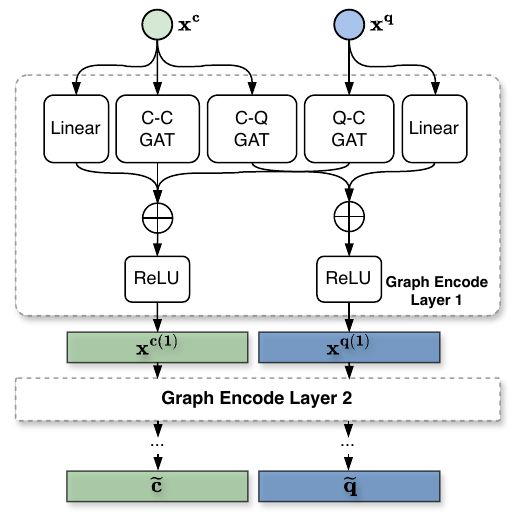}
    \caption{Structural Information Encoder (SIEnc).}
    \end{subfigure}
    \begin{subfigure}{.35\textwidth}
        \includegraphics[width=0.9\linewidth]{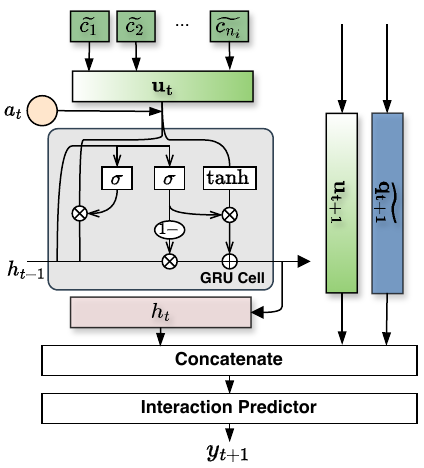}
        \caption{Student State Encoder and Response Predictor.}
    \end{subfigure}
    \caption{The framework and details of SINKT.}
    \label{fig:framework}
\end{figure*}

In this section, we will introduce our method in detail, and the overall framework is shown in Figure \ref{fig:framework}.
After generating the concept-question graph by GPT-4, we first use the Textual Information Encoder (TIEnc) to extract semantic information from the concepts and questions.
Then, we design a Structural Information Encoder (SIEnc) to learn question and concept representations from the heterogeneous graph. 
Finally, we design a student state encoder based on concept-level mastery to capture the student's knowledge state, and predicts the final response by an interaction predictor.
\subsection{Textual Information Encoder}
The introduction of LLMs bridges the gap between semantic information and plain text.
SINKT utilizes a Pretrained Language Model (PLM) as semantic encoder to get representations of the plain text of concepts and questions.
For each concept $c_i$, we input its plain text $\text{TEXT}(c_i)$ into the PLM and acquire its representation vector
\begin{equation}
\mathbf{x}_i^c = \text{PLM}_c(\text{TEXT}(c_i)) \in \mathbb R^{d_t},
\end{equation}
where $d_t$ is the encoding dimension of the PLM.
Similarly, for each question $q_i$, we could obtain its representation vector
\begin{equation}
\mathbf{x}_i^q = \text{PLM}_q(\text{TEXT}(q_i)) \in \mathbb R^{d_t}.
\end{equation}
The representation vectors $\mathbf{x}_i^c$ and $\mathbf{x}_i^q$ contain the semantic information of concepts and questions.
In our framework, we fix the parameters of PLM encoder and use representation vectors as an input of following modules.

\subsection{Structural Information Encoder}
When a student answers a question, their mastery of concepts associated with that question influence their response. The mastery of both the target concept and its related concepts, as well as the related questions, can provide valuable information for the assessment of KT models. Therefore, the three types of edges in the heterogeneous graph G, \textit{i.e.,} concept-question, concept-concept, and question-concept, are considered by SINKT. To encode such structural information, we carefully design a multi-layer heterogeneous graph encoder.

 In the graph $\mathcal G$, denote neighbor questions of the concept $c_i$ be $\mathcal N^c_{c_i}$ and $\mathcal N^q_{c_i}$, respectively.
 The set of neighbor concepts of the question $q_i$ is denoted as $\mathcal N^c_{q_i}$.
 We apply three different Graph Attention Networks (GAT) to aggregate neighborhood information for vertexes.
 Concept-Question GAT layer fuse neighbor concept representations of the target concept $c_i$. Specifically, the Concept-Concept GAT can be expressed as:
\begin{equation}
     \alpha^{cq}_{i,j} = \frac{\exp \left(\text{LeakyReLU}\left(\mathbf{a_{cc}}^T (\mathbf{x}^q_i\oplus \mathbf{x}^c_j)\right)\right)}{\sum_{c_k \in \mathcal{N}^q_{c_i}} \exp \left(\text{LeakyReLU}\left(\mathbf{a_{cc}}^T (\mathbf{x}^q_i\oplus \mathbf{x}^c_k)\right)\right)},
 \end{equation}
  \begin{equation}
     \mathbf{e}^{cq}_i = \sum_{c_j \in \mathcal{N}^q_{c_i}} \alpha^{cq}_{i,j}  * \left(\mathbf{W}_{cq}\mathbf{x}^{c}_j\right),
\end{equation}
where $\mathbf{W_{cq}}$ is the aggregate weight $\mathbf a_{cq}$ is the attention weight, $\oplus$ denotes concatenate operation.
Similarly, Concept-Concept GAT integrate neighbor concept information to the target concept:
\begin{equation}
     \alpha^{cc}_{i,j} = \frac{\exp \left(\text{LeakyReLU}\left(\mathbf{a_{cc}}^T (\mathbf{x}^c_i\oplus \mathbf{x}^c_j)\right)\right)}{\sum_{c_k \in \mathcal{N}^c_{c_i}} \exp \left(\text{LeakyReLU}\left(\mathbf{a_{cc}}^T (\mathbf{x}^c_i\oplus \mathbf{x}^c_k)\right)\right)}, 
     \label{alpha}
 \end{equation}

 \begin{equation}
     \mathbf{e}^{cc}_i = \sum_{c_j \in \mathcal{N}^c_{c_i}} \alpha^{cc}_{i,j} * \left(\mathbf{W}_{cc}\mathbf{e}^{c}_j\right),
\end{equation}
and Question-Concept GAT integrate neighbor concept information to the target concept:
\begin{equation}
     \alpha^{qc}_{i,j} = \frac{\exp \left(\text{LeakyReLU}\left(\mathbf{a_{qc}}^T (\mathbf{x}^c_i\oplus \mathbf{x}^q_j)\right)\right)}{\sum_{q_k \in \mathcal{N}^c_{q_i}} \exp \left(\text{LeakyReLU}\left(\mathbf{a_{qc}}^T (\mathbf{x}^c_i\oplus \mathbf{x}^q_k)\right)\right)},
 \end{equation}
  \begin{equation}
     \mathbf{e}^{qc}_i = \sum_{q_j \in \mathcal{N}^c_{q_i}} \alpha^{qc}_{i,j} * \left(\mathbf{W}_{qc}\mathbf{x}^{q}_j\right).
\end{equation}
To emphasize the semantic information inherent in concepts and questions themselves, SINKT introduces jumping knowledge~\cite{junpingknowledge} to propagate the vertex original representations. The node representation of $l$-th encoder layer of can be represented as:
\begin{equation}
    \mathbf{x}^{c(l)}_i = \text{ReLU}\left( \mathbf{W_c} \mathbf{x}^{c(l-1)}+ \mathbf{e}_i^{cc} + \mathbf{e}_i^{qc}\right),
    \end{equation}
\begin{equation}
     \mathbf{x}^{q(l)}_i = \text{ReLU}\left( \mathbf{W_q} \mathbf{x}^{q(l-1)}+ \mathbf{e}_i^{cq}\right),
\end{equation}
where $\mathbf{W_c}$ and $\mathbf{W_q}$ are trainable weight matrix for concepts and questions.
We initialize the input of the SIEnc by the output of the semantic information encoder:
\begin{equation}
    \mathbf{x}_{i}^{q(0)} = \mathbf{x}_i^q, \mathbf{x}_{i}^{c(0)} = \mathbf{x}_i^c.
\end{equation} 
We use $\widetilde{\mathbf{q_i}} \in \mathbb R^d$ and $\widetilde{\mathbf{c_i}} \in \mathbb R^d$ to denote representation vectors of the question $q_i$ and the concept $c_i$ after the $k$-layer graph encoder, where $d$, $k$ are hyper-parameters.

\subsection{Student State Encoder}
To effectively learn from highly sparse question-response data,  \cite{akt,liu2023pykt, liu2023simplekt} suggest to transform the original question-response data into concept-response data.
Due to the fact that some questions contain multiple concepts, SINKT averages representation vectors of concepts related to the question $q_t$ to represent students' learning at time step $t$
\begin{equation}
    \mathbf{u}_t = \frac{1}{|\mathcal  C_{q_t}|}\sum_{c_i\in \mathcal C_{q_t}}\widetilde{\mathbf{c}_i}.
\end{equation}
To jointly represent the item and correctness of students' response, we introduce $\mathbf{v}_t\in \mathbb{R}^{2d}$ as
\begin{equation}
\mathbf{v}_t=\left\{
\begin{array}{ll}
\mathbf{u}_t \oplus \mathbf{0} & r_t=1 \\
\mathbf{0} \oplus \mathbf{u}_t & r_t=0
\end{array},\right.
\end{equation}
where $\mathbf{0}\in\mathbb{R}^d$ is a zero-vector.
To model learning history of students, we use Gated Recurrent Unit (GRU)~\cite{GRU} to capture the sequential learning behavior
\begin{align}
    \mathbf{u}_r &= \sigma(\mathbf{W}_r (\mathbf{v}_i \oplus \mathbf{h}_{t-1} ) + \mathbf{b}_r,\\
    \mathbf{u}_z &= \sigma(\mathbf{W}_z  (\mathbf{v}_i \oplus \mathbf{h}_t) + \mathbf{b}_z,\\
    \mathbf{u}_h &= \tanh(\mathbf{W}_h  (\mathbf{v}_i \oplus (\mathbf{u}_r * \mathbf{h}_{t-1})) + \mathbf{b}_h),\\
    \mathbf{h}_t &=  (1-\mathbf{u}_z) * \mathbf{u}_h + \mathbf{u}_z * \mathbf{h}_{t-1}.
\end{align}
Here $\sigma$ denotes the \textit{sigmoid} function, and  $\mathbf{W}_r, \mathbf{b}_r,\mathbf{W}_z, \mathbf{b}_z,\mathbf{W}_h, \mathbf{b}_h$ are trainable parameters, where $\mathbf{W}_r,\mathbf{W}_z, \mathbf{W}_h \in \mathbb R^{d\times 2d}$ and $\mathbf{b}_r,\mathbf{b}_z, \mathbf{b}_h \in \mathbb R^{d}.$
Hidden state $\mathbf{h}_t \in \mathbb{R}^{d}$ represents the knowledge state of the student at time step $t$.
The student state encoder considers the concept-level forgetting and knowledge acquisition pattern together during student's learning process.

\subsection{Response Prediction}
When a student answers a question, his knowledge state of the corresponding concepts and the description of the question would influence his correctness.
Given knowledge state of the student $\mathbf{h_t}$ at time step $t$, the question representation vector $\mathbf{\widetilde{q}_{t+1}}$ and its concept-level representation vector $\mathbf{u_{t+1}}$, we make the prediction as follows
\begin{equation}
{y}_{t+1} = \sigma\left(\mathbf{W_p}\left(\mathbf{h}_t\oplus \mathbf{\tilde{q}}_{t+1} \oplus \mathbf{u}_{t+1}\right) + \mathbf{b}_p\right).
\end{equation}
where $\mathbf{W}_p \in \mathbb R^{3d}$ and $\mathbf{b}_p\in \mathbb R$ are trainable parameters.
To train all parameters in SINKT, we choose the cross-entropy log loss between the predicted response $ y_{t+1}$ and the ground-truth response $r_{t+1}$ as the objective function
\begin{equation}
    \mathcal L = -\sum_{t=1}^T\left( r_t \log y_t + (1-r_t)\log (1-y_t) \right).
\end{equation}

\subsection{Automated Pipeline for Inductive KT}
\begin{figure}[t]
    \centering
    \includegraphics[width=0.9\linewidth]{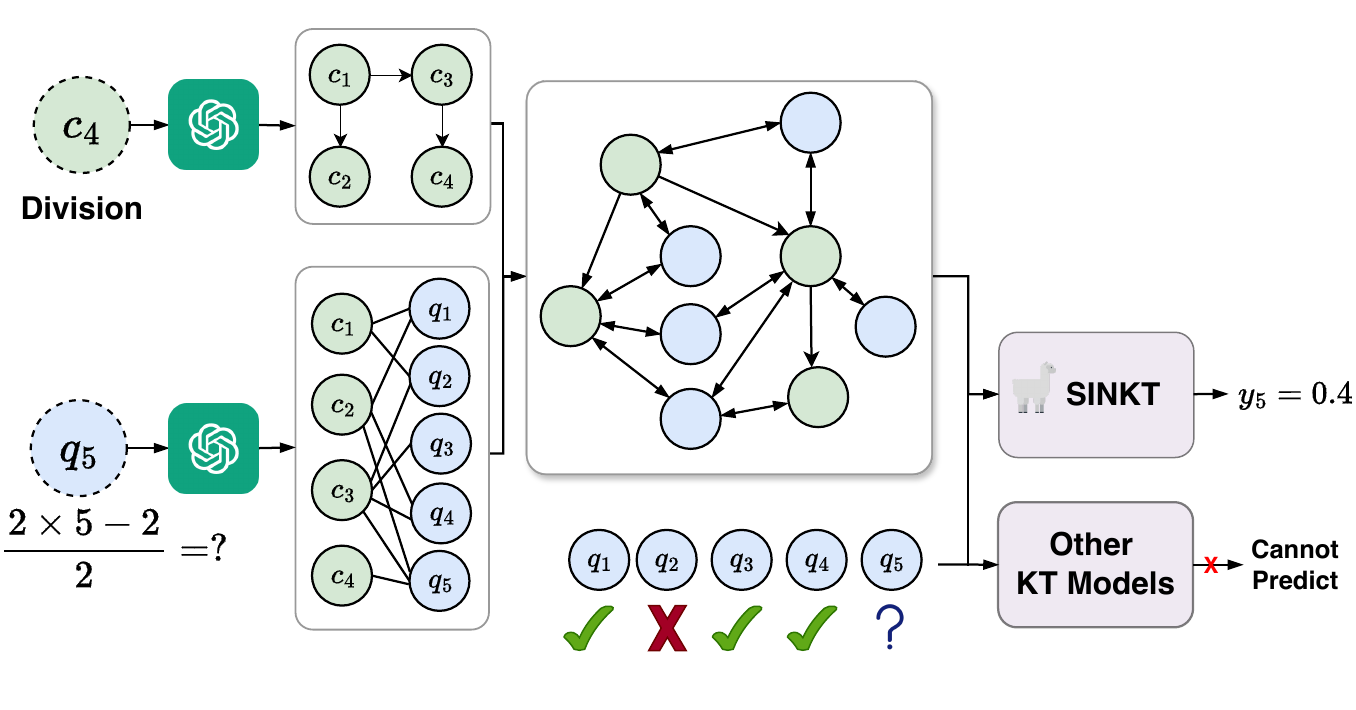} 
    \vspace{-20pt}
    \caption{An automated pipeline for the inductive KT task. In the graph, a new concept $c_4$ and a new question $q_5$ are added into the database. }
    \vspace{-10pt}
    \label{fig:pipeline}
\end{figure}

SINKT mainly focuses on the semantic and structural information, which could be completely generated by LLMs.
Meanwhile, SINKT use textual representations instead of ID embeddings, which is available even in the absence of learning history.
Therefore, when a new concept or a new question is introduced into the database of ITS, SINKT is capable to capture the knowledge state of students towards the new questions or concepts through automated processing. 
Figure~\ref{fig:pipeline} illustrates the automated ingestion process for $c_4$ and $q_5$ in Figure~\ref{fig:intro}. 
When a teacher adds a new concept to the ITS, LLMs are invoked to integrate the new vertex into the Concept-Question Graph.
The teacher could continue to add a new question that is related to this new concept, and the ITS further utilizes LLMs to annotate the new question with concepts. 
The new concept-question heterogeneous graph is automatically constructed.
When it is necessary to assess a student's mastery  of the new question $q_5$, SINKT can automatically perform text encoding and propagate information through the graph, ultimately accomplishing response prediction. 

The automated ingestion pipeline enhances the ITS by providing the flexibility to update and expand the database. This capability not only streamlines the process of integrating new content but also ensures that the system remains adaptive to evolving educational needs and emerging topics.
However, existing transductive KT models are unable to achieve this.

\section{Experiment}
In this section, we conduct experiments on four real-world datasets to evaluate the proposed SINKT framework. Specifically, we aim to answer the following research questions:
\begin{itemize}
    \item[\textbf{RQ1}] How does the proposed SINKT performs compared to the state-of-the-art KT models?
    \item[\textbf{RQ2}] How does SINKT perform when a new question is added to the ITS?
    \item[\textbf{RQ3}] What is the influence of various componets of SINKT?
    \item[\textbf{RQ4}]  How do different hyperparameters affect the performance of SINKT?
    \item[\textbf{RQ5}] Is the concept relation graph generated by GPT-4  reasonable and useful?
\end{itemize}
\subsection{Datasets}
In our experiments, we evaluate our method on four datasets:
(1) ASSIST09\footnote{https://sites.google.com/site/assistmentsdata/home/2009-2010-assistment-data?authuser=0}, (2)ASSIST12\footnote{https://sites.google.com/site/ASSISTdata/home/2012-13-school-data-with-affect}, (3) Junyi\footnote{https://pslcdatashop.web.cmu.edu/DatasetInfo?datasetId=1198} and (4) Programming.
ASSIST09, ASSIST12, and Junyi are public datasets widely used in intelligent education research.
Programming is a private dataset with plain text of questions and concepts.
For each dataset, students answering less than ten questions are removed. 
In order to simulate the situation of data sparsity, we choose 2,000 students' learning history in ASSIST12 and Junyi.
The basic statistics of these four datasets are listed in Table~\ref{tab:dataset_statistics}, and descriptions are as follows:
\begin{itemize}[leftmargin=10pt]
    \item \textbf{ASSIST09} is collected from an online tutor platform that teaches students mathematics during the school year 2009 to 2010. The dataset records the plain text of concepts and does not describe questions.

    \item \textbf{ASSIST12} is also collected from the ASSIST platform from Sept 2012 to Oct 2013. Like ASSIST09,  plain text of concepts is contained, but question descriptions are not provided.
    \item \textbf{Junyi} is collected between November 2010 and March 2015 from the e-learning platform in Taiwan Junyi Academy. We choose KC as the concept text without using question descriptions.
    \item \textbf{Programming} is collected from a commercial code-learning platform from December 13, 2021 to February 17, 2023. The dataset contains plain text of concepts and question descriptions.
\end{itemize}

\renewcommand{\arraystretch}{1.0} %
\begin{table}[t]
\caption{Statistics of all datasets. Q. and C. denote questions and concepts. Edge density is the density of the concept correlation graph generated by GPT-4.}
\centering
\label{tab:dataset_statistics}
\resizebox{\linewidth}{!}{%
\begin{tabular}{l|c|c|c|c}
\hline\toprule
\textbf{Statistics} & \textbf{ASSIST09} & \textbf{ASSIST12} & \textbf{Junyi} & \textbf{Programming} \\
\midrule
Subjects & Math &Math & Math& Programming\\
Students       & 2,661   & 2,000  & 2,000 &2,756 \\
Interactions      & 165,455     & 159,545 & 93,697 & 193,284    \\
Questions     & 14,083     & 31,229    & 2,017 & 726     \\
Concepts      & 134      & 240      & 40 & 82      \\ \midrule
Q. per C. &  105.10   & 130.12    & 50.43  & 17.89 \\
Attempts per Q.& 26.04  & 7.70  & 151.86  & 603.50\\
 Edge Density &0.1067 & 0.0778 & 0.1750 & 0.0723\\
\bottomrule\hline
\end{tabular}
}
\vspace{-15pt}
\end{table}

\subsection{Baselines}
To demonstrate the effectiveness of predicting students' response, we evaluate the performance of SINKT with 12 baselines. 
\begin{itemize}[leftmargin=10pt]
    \item \textbf{DKT}~\cite{dkt} and \textbf{DHKT}~\cite{dhkt} use deep learning methods on the KT task, which uses RNN to model students' learning history.
    \item \textbf{DKVMN}~\cite{dkvmn}, \textbf{SKVMN}~\cite{skvmn} apply key-value memory network on the KT task.
    \item\textbf{CKT}~\cite{ckt} applies hierarchical convolutional layers and learns a matrix demonstrating students' mastery level of each concept. 
    \item \textbf{SAKT}~\cite{sakt}, \textbf{EERNNA}~\cite{eernna} and \textbf{AKT}~\cite{akt} use attention mechanism to learn the past interactions' importance in predicting students' current questions.
    \item \textbf{GKT}~\cite{gkt} generates a transition graph from the dataset and uses GNN to encode the knowledge states of students.
    \item \textbf{SKT}~\cite{skt} exploits similarity relation and prerequisites relation between concepts based on GKT.
    \item \textbf{IEKT}~\cite{iekt} retrieves the similar histories of other students to enhance the prediction of target students.
    \item \textbf{LBKT}~\cite{lbkt} considers several dominant learning behaviors, including speed, attempts, and hints in the learning process.
\end{itemize}
\begin{table*}[t]
\caption{Overall performance of SINKT and baselines in four real-world datasets. SINKT-BERT and SINKT-Vicuna choose BERT and Vicuna as TIEnc, respectively. Existing state-of-the-art results are underlined and the best results are bold. * indicates p-value $< 0.05$ in the t-test. }
\centering
\setlength{\tabcolsep}{8pt} 
\label{tab:rq1}
\begin{tabular}{@{\hspace{5mm}}c|l|cc|cc|cc|cc@{}}
\hline\toprule
\multicolumn{1}{c|}{\multirow{2}{*}{{\textbf{Groups}}}} & \multicolumn{1}{c|}{\multirow{2}{*}{{\textbf{Models}}}} & \multicolumn{2}{c|}{\textbf{ASSIST09}} & \multicolumn{2}{c|}{\textbf{ASSIST12}} & \multicolumn{2}{c|}{\textbf{Junyi}} & \multicolumn{2}{c}{\textbf{Programming}} \\ 
\cmidrule(r){3-4} \cmidrule(lr){5-6} \cmidrule(lr){7-8} \cmidrule(l){9-10}
  \multicolumn{1}{c|}{}&\multicolumn{1}{c|}{}& \multicolumn{1}{c}{\textbf{ACC}} & \multicolumn{1}{c|}{\textbf{AUC}}  & \multicolumn{1}{c}{\textbf{ACC}} &  \multicolumn{1}{c|}{\textbf{AUC}}  & \multicolumn{1}{c}{\textbf{ACC}}&  \multicolumn{1}{c|}{\textbf{AUC}}  & \multicolumn{1}{c}{\textbf{ACC}} &  \multicolumn{1}{c}{\textbf{AUC}} \\ \midrule

\multirow{9}{*}{ID Only}& \multicolumn{1}{l|}{DKT}  &          0.7166& 0.7087                &0.6936 &0.6235 & 0.7516 & 0.7768   &   0.7799& 0.7957             \\
&\multicolumn{1}{l|}{DHKT}              & 0.7357        & 0.7442       & 0.7038        & 0.6627       & 0.7563      & 0.7826      & 0.7891      & 0.8130       \\
&\multicolumn{1}{l|}{DKVMN }            & 0.7175        & 0.7033       & 0.6975        & 0.6282       & 0.7625      & 0.7949      & 0.7655      & 0.7651      \\
&\multicolumn{1}{l|}{SKVMN}             & 0.7201        & 0.7322       & 0.6952     & 0.6252    & \underline{0.7675}   & \underline{0.8018}   & 0.7687     & 0.7736      \\
&\multicolumn{1}{l|}{CKT}               & 0.6672         & 0.6725       & 0.6492        & 0.5722       & 0.7268      & 0.7446      & 0.7721      & 0.7815      \\
& \multicolumn{1}{l|}{SAKT}              & 0.6509        & 0.6370       & 0.6626        & 0.5557       & 0.7381      & 0.7580      & 0.7558      & 0.7680      \\
&\multicolumn{1}{l|}{EERNNA}       & 0.7247        & 0.7427       & 0.6964        & 0.6487       & 0.7558      & 0.7837      & 0.7867      & 0.8071      \\
&\multicolumn{1}{l|}{AKT}               & 0.6683     & 0.6158    & 0.6652        & 0.5983       & 0.6864      & 0.6753      & 0.7308      & 0.6251     \\
&\multicolumn{1}{l|}{SKT}              & 0.7082     & 0.7157    & 0.6858        & 0.6165       & 0.7307      & 0.7335      & 0.7781       & 0.7898      \\\midrule
\multirow{3}{*}{Using Extra Info.} &\multicolumn{1}{l|}{GKT}               & \underline{0.7370}     & \underline{0.7511}    & \underline{0.7160}        & \underline{0.6918}       & 0.7345      & 0.7298      & 0.7641        & 0.7507       \\
&\multicolumn{1}{l|}{IEKT}  & 0.7031 & 0.6966 & 0.6880 & 0.6216 & 0.7168 & 0.7081 & \underline{0.7973}  &\underline{0.8261}\\
&\multicolumn{1}{l|}{LBKT}              & 0.6819     & 0.6814    & 0.6747     & 0.6707     & 0.7456   & 0.7733      & 0.7780       & 0.7898   \\ \midrule
\multirow{2}{*}{Ours}&\multicolumn{1}{l|}{SINKT-BERT} & \textbf{0.7416}* & \textbf{0.7726}*& \textbf{0.7264}*&	\textbf{0.7028}*&	\textbf{0.7721}* &	0.8069 &0.7957 &	0.8225\\
&\multicolumn{1}{l|}{SINKT-Vicuna} & 0.7417& 0.7700&0.7173	&0.7003 & 0.7718 & \textbf{0.8095}* &	\textbf{0.7980}*&	\textbf{0.8267}* \\
\bottomrule\hline
\end{tabular}
\end{table*}

\subsection{Implementation Details}
In our experiment, we choose the most recent 200 records of students.
We choose BERT to be the textual encoder for three public datasets with concept text and choose Vicuna to be the textual encoder for the Programming dataset with concept and question text. 
We will compare the performance of different textual encoders in Section \ref{sec:rq1}.
The layer number of the graph encoder is chosen from $\{1,2,3\}$.
The parameter $d$ is set to $256$.
The learning rate is chosen from $\{0.0001, 0.00005\}$ with a decay at each epoch. The optimizer is Adam~\cite{Kingma2014AdamAM}.
For a fair comparison, the hyper-parameters of baselines are carefully chosen to have the best performance.
Our code is available now\footnote{The PyTorch and MindSpore implementation is available at: https://github.com/tubehao/SINKT.git and https://github.com/mindspore-lab/models/tree/master/research/huawei-noah/SINKT.}.

\subsection{Overall Performance (RQ1)}\label{sec:rq1}

We compare SINKT to 12 baselines on predicting the correctness of students' responses in four real-world datasets.
The experimental results are shown in Table \ref{tab:rq1}.
We use Accuracy (ACC) and Area Under the Curve (AUC) as the evaluation metric.

In Table \ref{tab:rq1}, several observations can be obtained as follows:
\begin{enumerate}[leftmargin=10pt]
    \item SINKT significantly outperforms all comparative methods across all datasets and evaluation metrics. This superior performance indicates that SINKT provides more precise representations of concepts and questions, as well as a more accurate estimation of students' knowledge states. 
    \item When the datasets only contain textual information of concepts (ASSIST09, ASSIST12, Junyi), using BERT~\cite{devlin2018bert} as the textual encoder yields better results. This is attributed to the fact that most concept texts consist of 1-3 words, and models like BERT are more adept at encoding word-level text. In contrast, when SINKT is provided with question texts (Prgogramming), Vicuna~\cite{vicuna2023} demonstrates improved performance, since generative models have a better understanding of sentence-level text typical of question descriptions.
    \item Introducing additional specific information into the model does not guarantee beneficial results across all datasets. For example, GKT, which extracts a transition graph from training data to train concept embeddings, shows significant performance improvement on the ASSIST09 and ASSIST12 datasets. However, this approach introduces noise in the Junyi and Programming datasets, detracting from model effectiveness.
\end{enumerate}

\begin{table}[t]
\centering
\caption{Performance comparison of DHKT, EERNNA, and SINKT in transductive and inductive KT tasks on Programming dataset. }
\label{tab:rq2}
\begin{tabular}{c|lccc}
\hline\toprule
\multirow{2}{*}{\textbf{KT Task Type}}     & \multirow{2}{*}{\textbf{Metrics}} & \multicolumn{3}{|c}{\textbf{Models}}                                         \\ \cline{3-5} 
                                 &                          & \multicolumn{1}{|c}{\textbf{DHKT}}   & \multicolumn{1}{c}{\textbf{EERNNA}} & \textbf{SINKT}  \\ \midrule
{\multirow{2}{*}{Transductive KT}}& ACC                      & \multicolumn{1}{|c}{0.7890} & \multicolumn{1}{c}{0.7867} & 0.7957 \\ \cline{2-5}
                                 & AUC                      & \multicolumn{1}{|c}{0.8130} & \multicolumn{1}{c}{0.8071} & 0.8225 \\ \midrule
\multirow{2}{*}{Inductive KT}    & ACC                      & \multicolumn{1}{|c}{0.5931} & \multicolumn{1}{c}{0.5781} & 0.6497 \\ \cline{2-5} 
                                 & AUC                      & \multicolumn{1}{|c}{0.5442} & \multicolumn{1}{c}{0.5502} & 0.6071 \\ \bottomrule\hline
\end{tabular}
\end{table}

\begin{table*}[t]
\caption{Ablation study of SINKT on all datasets.}
\centering
\setlength{\tabcolsep}{8pt} 
\label{tab:rq4}
\begin{tabular}{@{\hspace{5mm}}l|cc|cc|cc|cc@{}}
\hline\toprule
\multicolumn{1}{c|}{\multirow{2}{*}{\textbf{Models}}} & \multicolumn{2}{c|}{\textbf{ASSIST09}} & \multicolumn{2}{c|}{\textbf{ASSIST12}} & \multicolumn{2}{c|}{\textbf{Junyi}} & \multicolumn{2}{c}{\textbf{Programming}} \\ 
\cmidrule(r){2-3} \cmidrule(lr){4-5} \cmidrule(lr){6-7} \cmidrule(l){8-9}
  & \multicolumn{1}{c}{\textbf{ACC}} & \textbf{AUC}  & \textbf{ACC} & \textbf{AUC}  & \textbf{ACC} & \textbf{AUC}  & \textbf{ACC}& \textbf{AUC} \\ \midrule
SINKT & {0.7416} & {0.7726}& {0.7264}&	{0.7028}&	{0.7721}* &  {0.8095} &	{0.7980}&	{0.8267}\\ \midrule
SINKT-Linear &0.7381&	0.7556	&0.7130&	0.6866& 0.7366 & 0.7395 &0.7755&	0.7842\\
SINKT-GAT & 0.7427	&0.7680	&0.7116&	0.6914	&0.7692&0.8034 &	0.7949 &	0.8224\\\midrule
SINKT-Text  &0.7375	&0.7706&	0.7141&	0.7004	&0.7666&	0.8023&	0.7947&	0.8191\\
SINKT-Transition & 0.7449	&0.7709 &0.7106 & 0.7006& 0.7700 & 0.8064&0.7971 & 0.8256\\
\bottomrule\hline
\end{tabular}
\end{table*}

\begin{figure}[t]
    \centering
  \begin{subfigure}{.48\linewidth}
  \centering
  \includegraphics[width=0.9\linewidth]{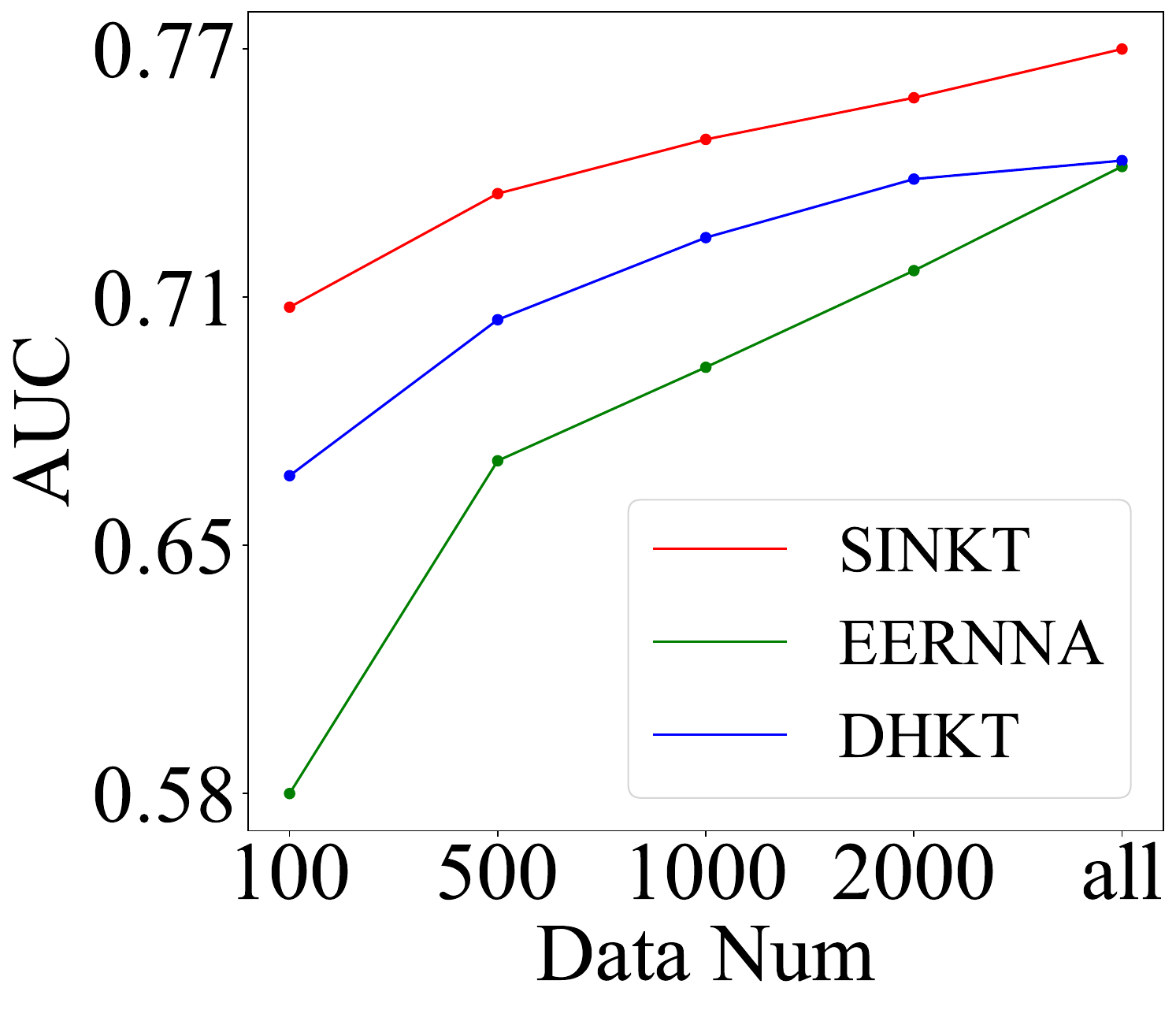}
    \caption{Effect of training samples.}
    \label{fig:sampleNum}
    \end{subfigure}
     \begin{subfigure}{.48\linewidth}
     \centering
  \includegraphics[width=0.9\linewidth]{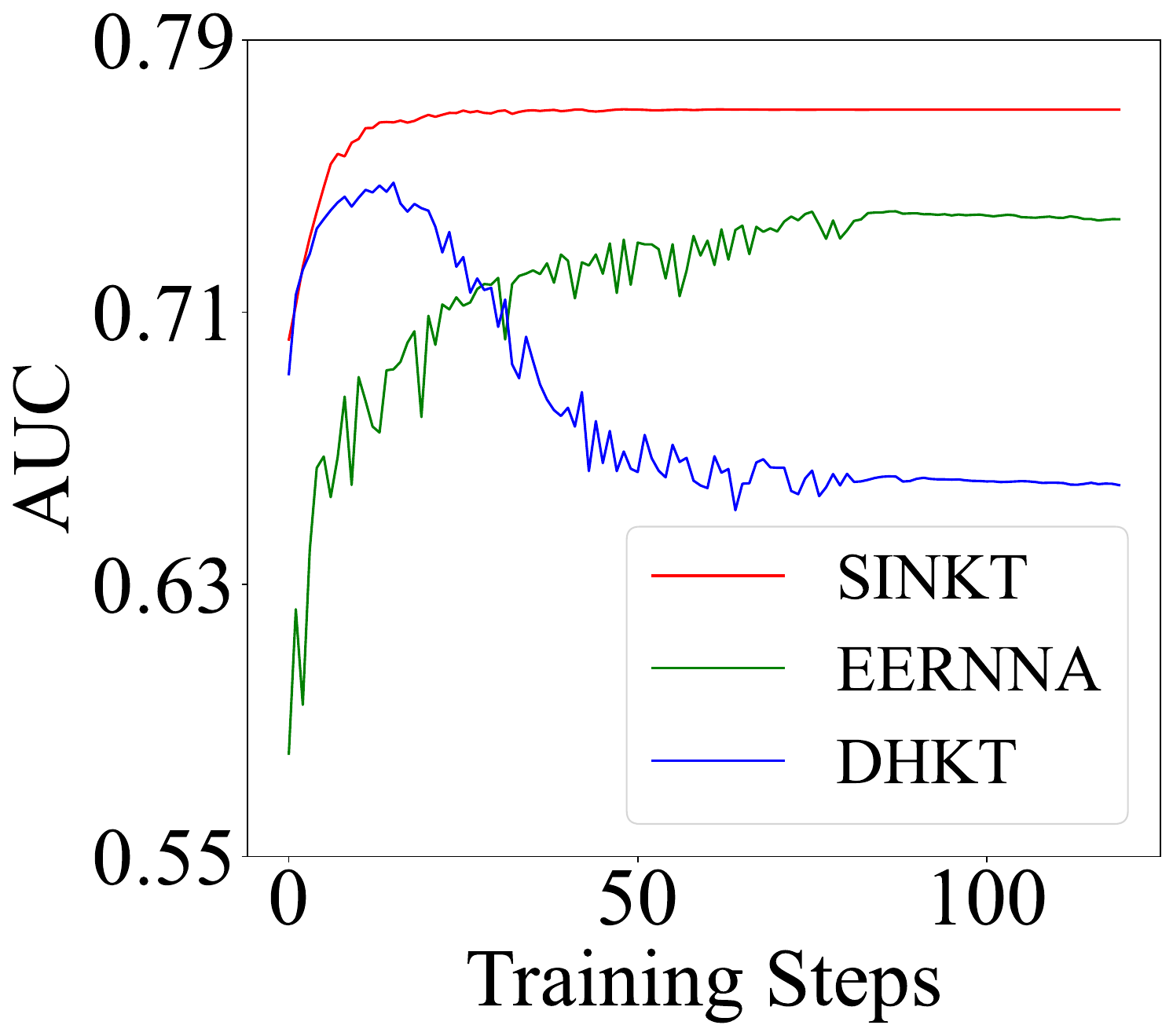}
    \caption{Rate of convergence.}
    \label{fig:trainCurve}
    \end{subfigure}
    \vspace{-5pt}
    \caption{Cold start analysis of SINKT.}
    \vspace{-5pt}
\end{figure}

\subsection{Inductive Learning of SINKT (RQ2)}
\begin{figure*}[t]
    \centering
    \begin{subfigure}{.24\linewidth}
        \centering
        \includegraphics[width=0.9\linewidth]{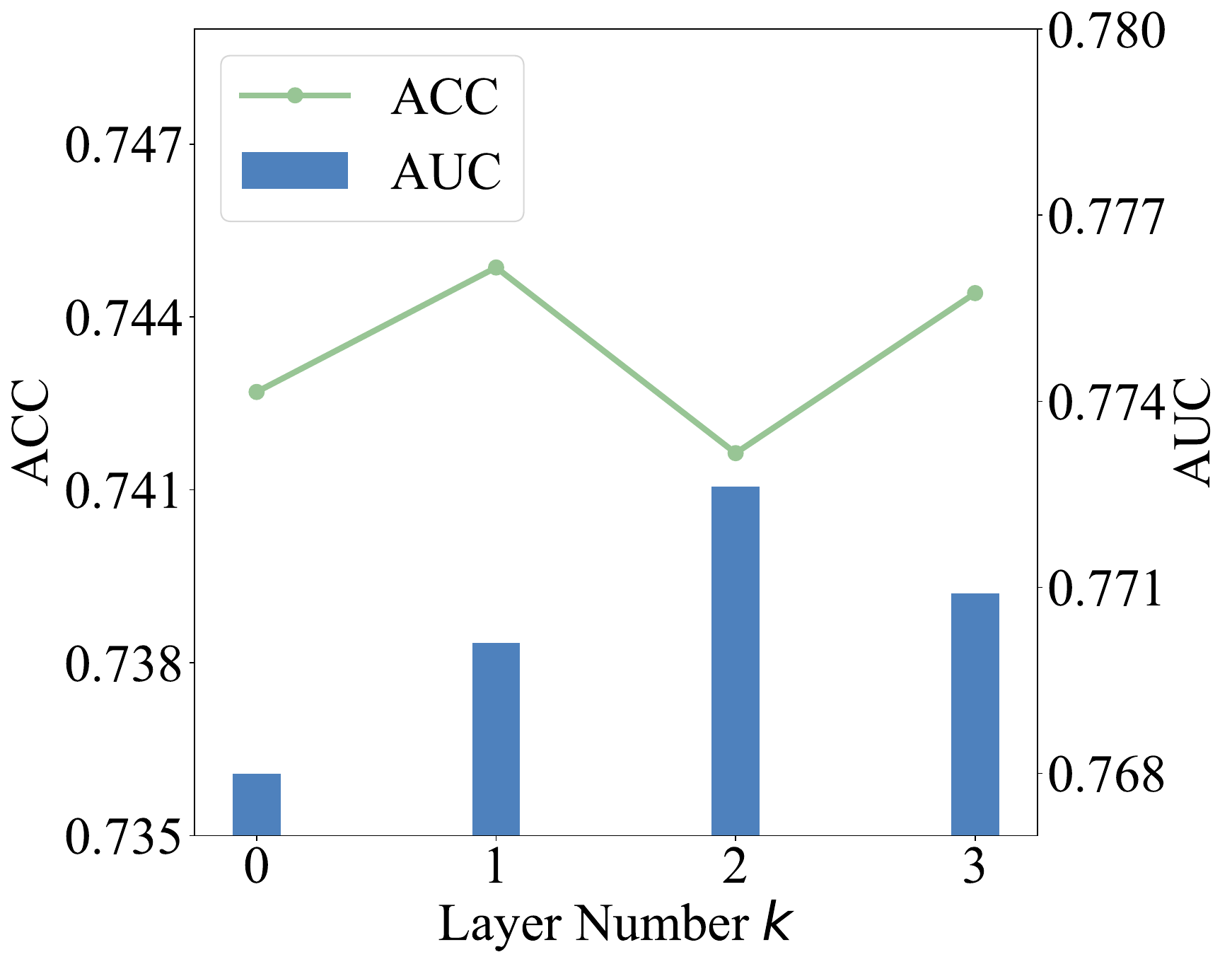}
        \caption{ASSIST09.}
    \end{subfigure}
    \begin{subfigure}{.24\linewidth}
        \centering
        \includegraphics[width=0.9\linewidth]{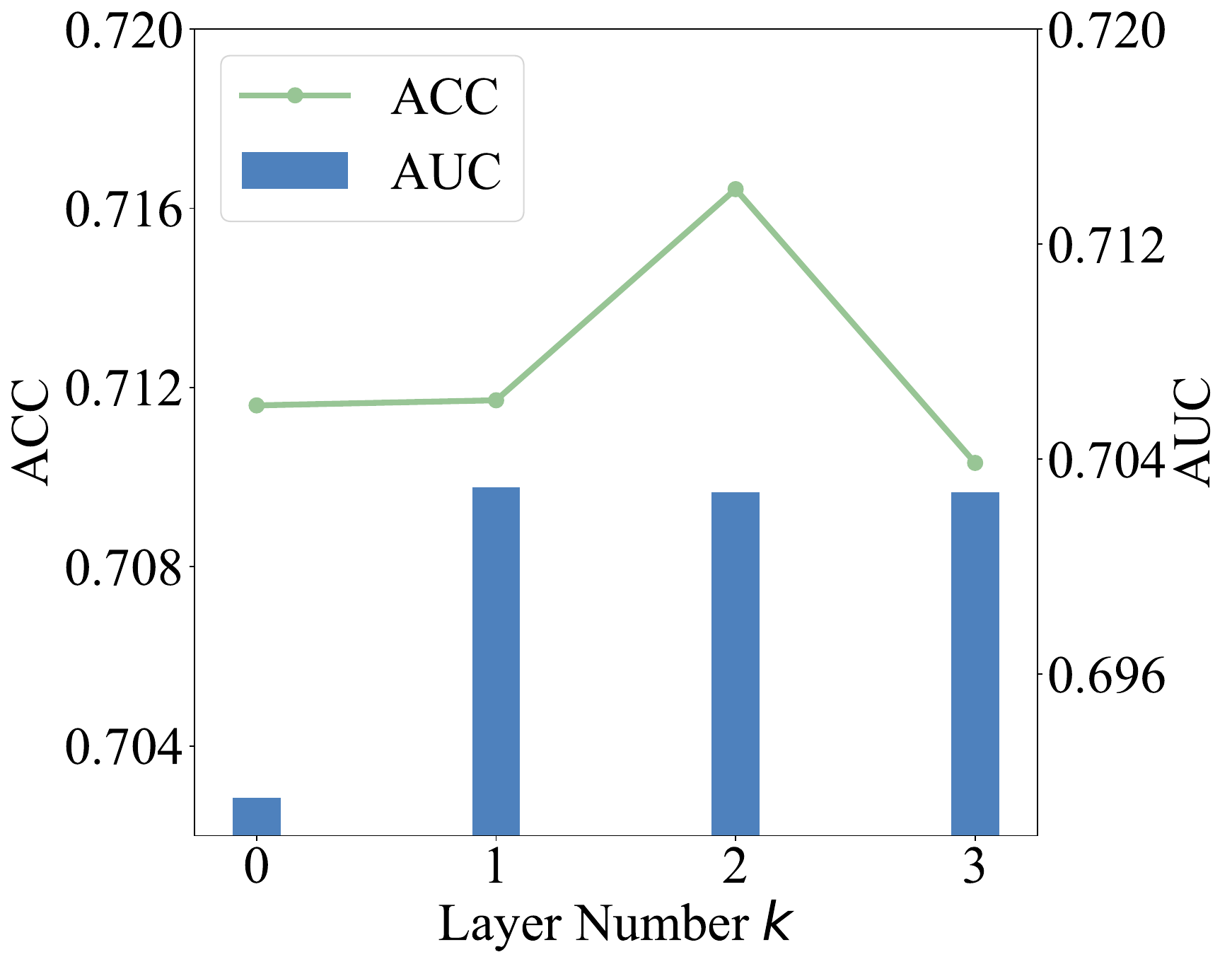}
        \caption{ASSIST12.}
    \end{subfigure}
    \begin{subfigure}{.24\linewidth}
        \centering
        \includegraphics[width=0.9\linewidth]{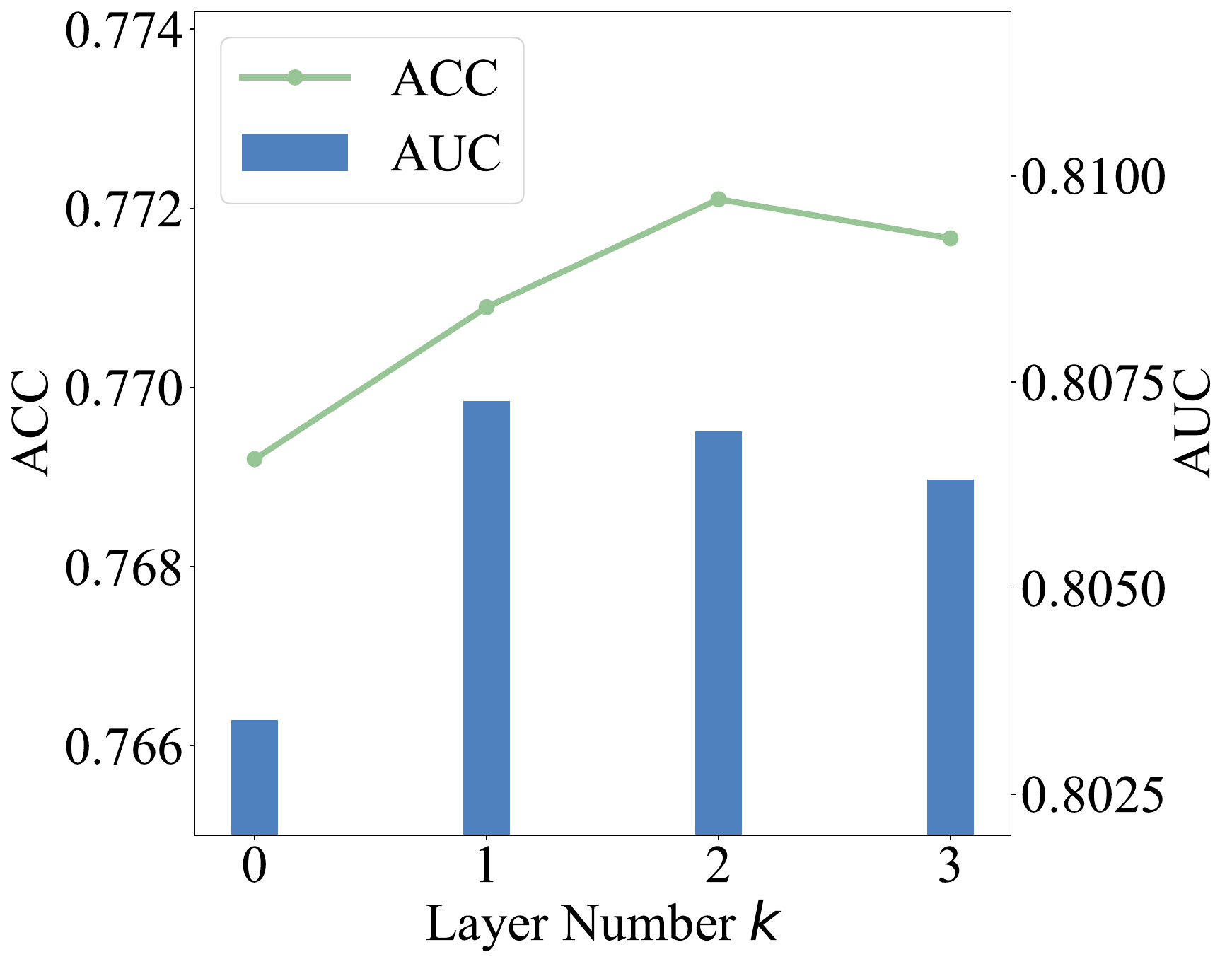}
        \caption{Junyi.}
    \end{subfigure}
    \begin{subfigure}{.24\linewidth}
        \centering
        \includegraphics[width=0.9\linewidth]{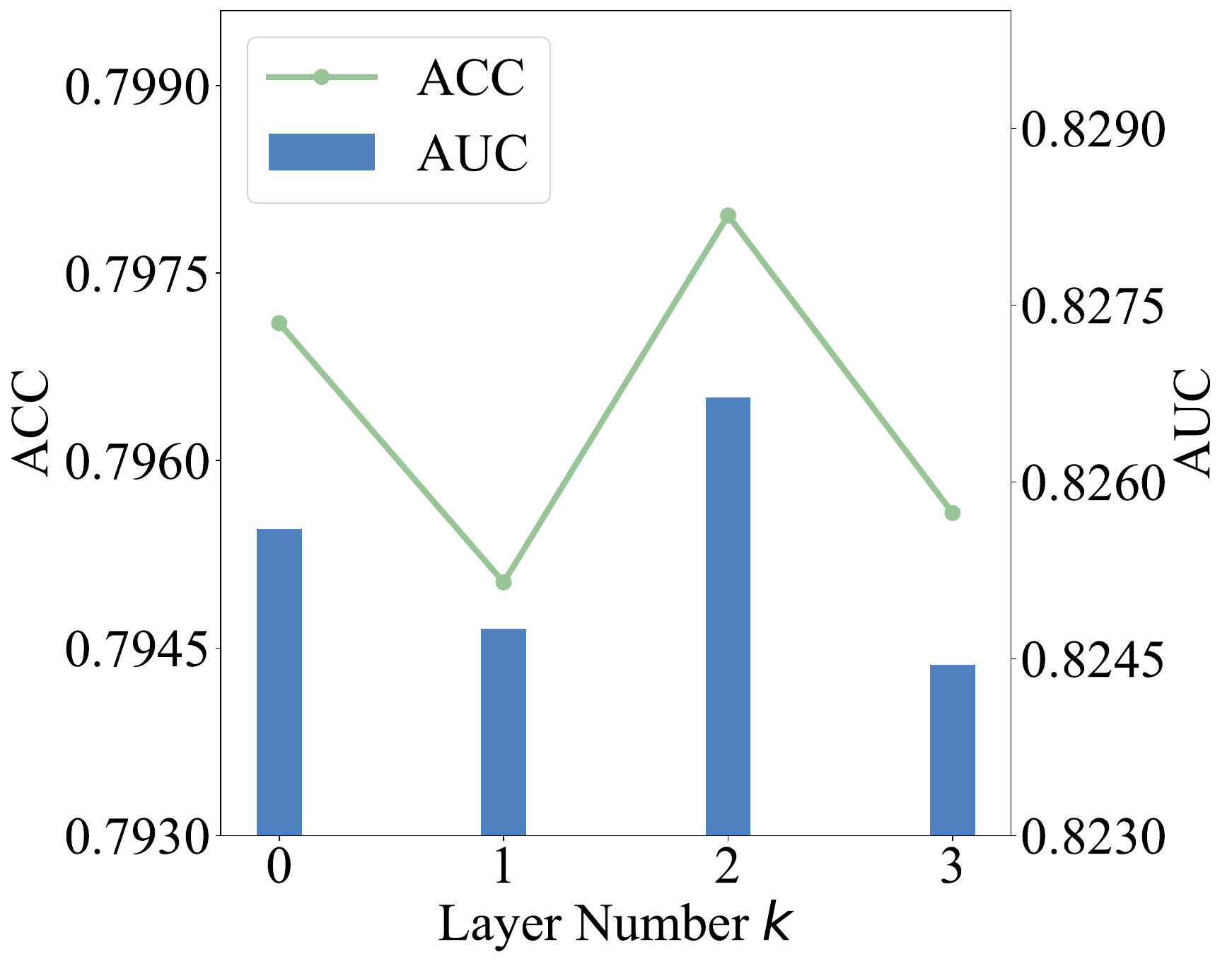}
        \caption{Programming.}
    \end{subfigure}
    \vspace{-5pt}
    \caption{Sensitivity analysis of layer number $k$ on four datasets.}
    \label{fig:sens_layernum}
\end{figure*}

\begin{figure*}[t]
    \centering
    \begin{subfigure}{.24\linewidth}
        \centering
        \includegraphics[width=0.9\linewidth]{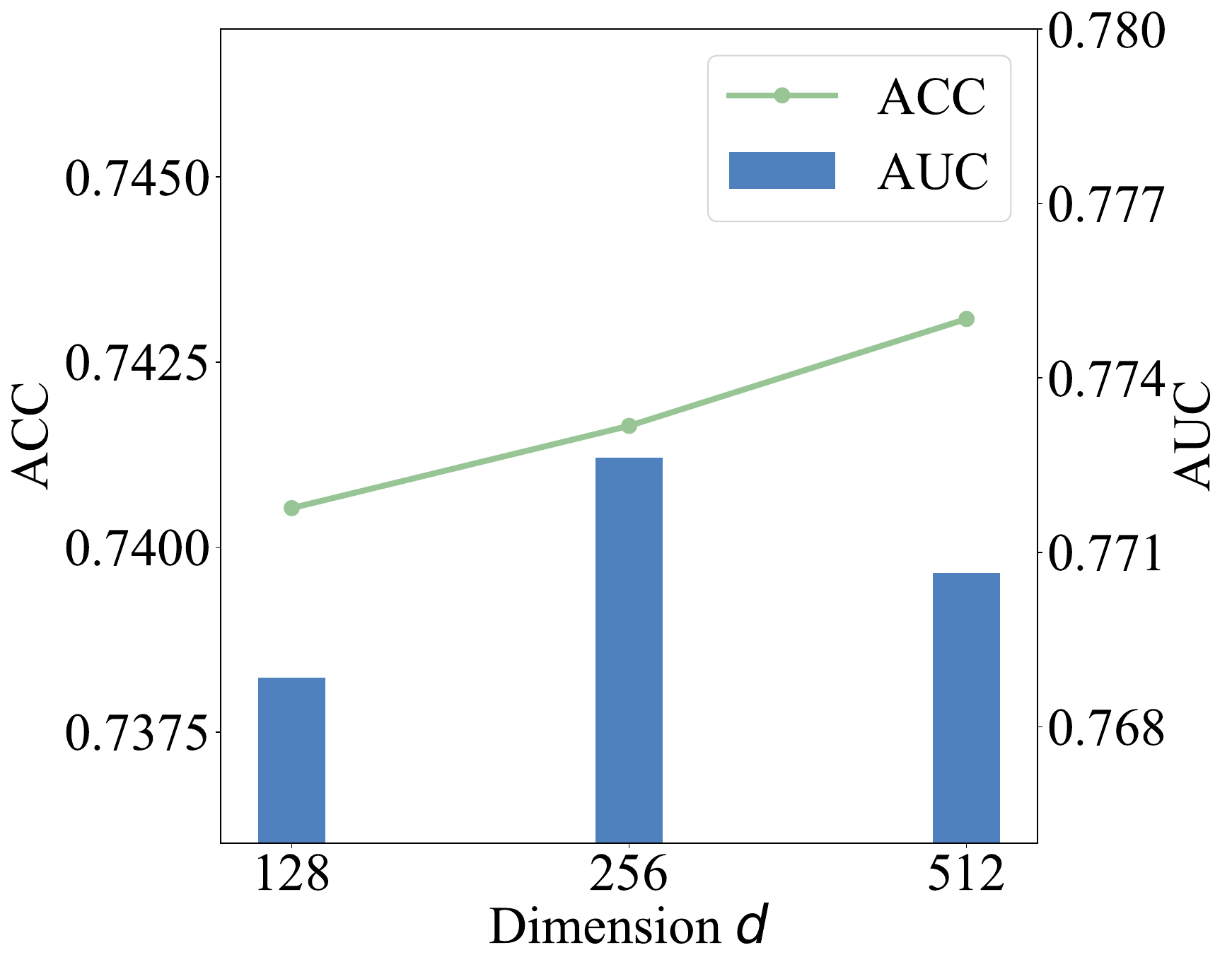}
        \caption{ASSIST09.}
    \end{subfigure}
    \begin{subfigure}{.24\linewidth}
        \centering
        \includegraphics[width=0.9\linewidth]{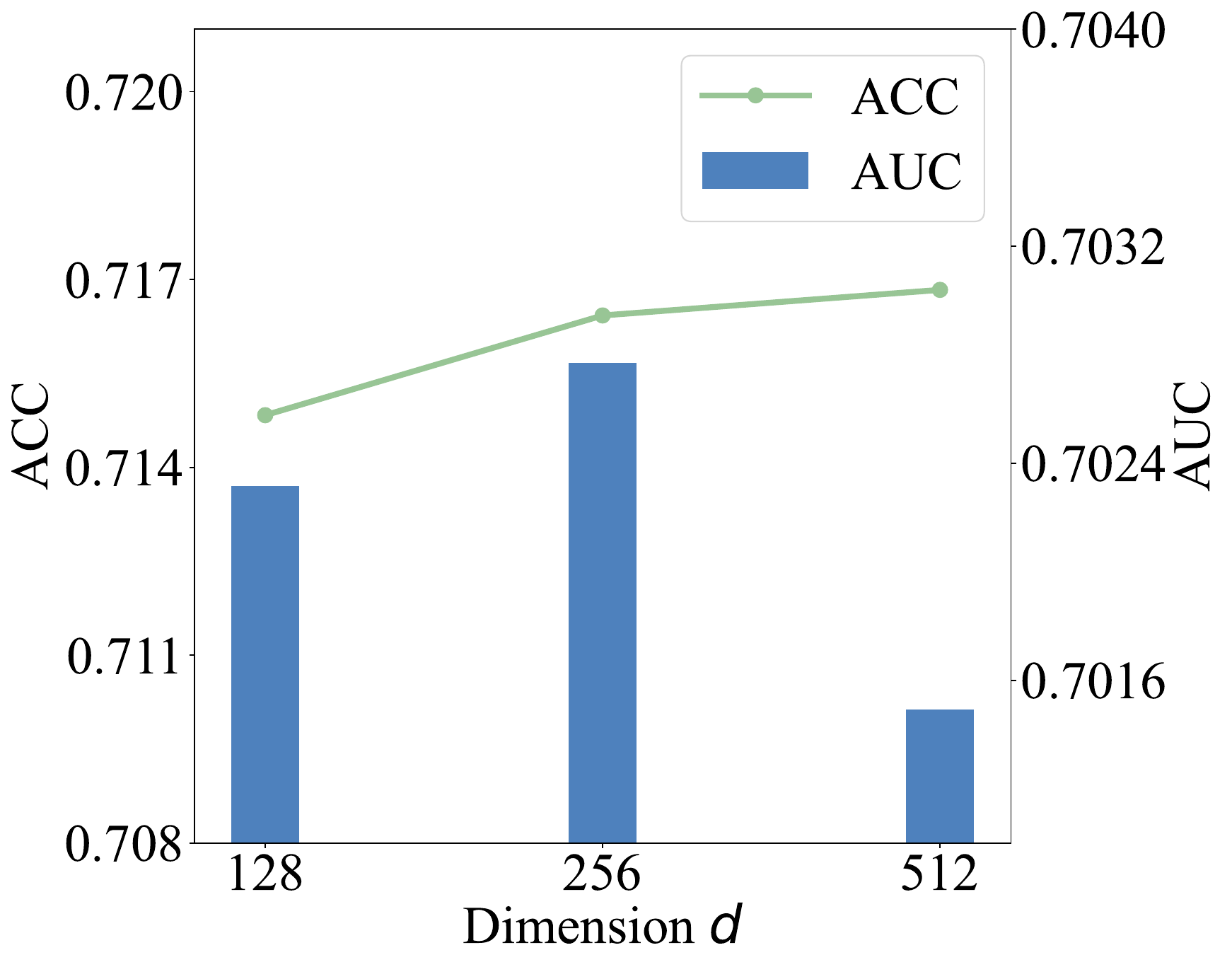}
        \caption{ASSIST12.}
    \end{subfigure}
    \begin{subfigure}{.24\linewidth}
        \centering
        \includegraphics[width=0.9\linewidth]{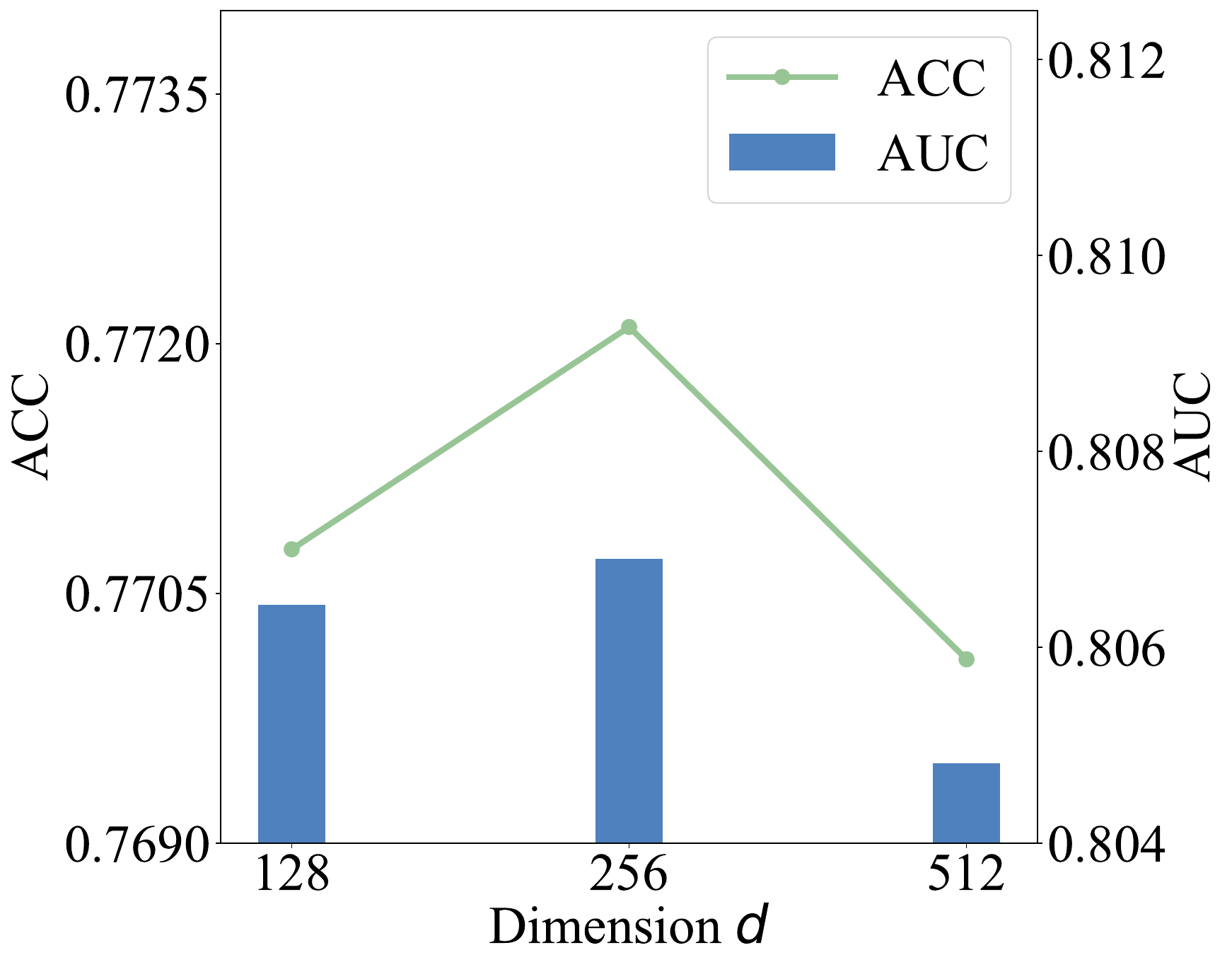}
        \caption{Junyi.}
    \end{subfigure}
    \begin{subfigure}{.24\linewidth}
        \centering
        \includegraphics[width=0.9\linewidth]{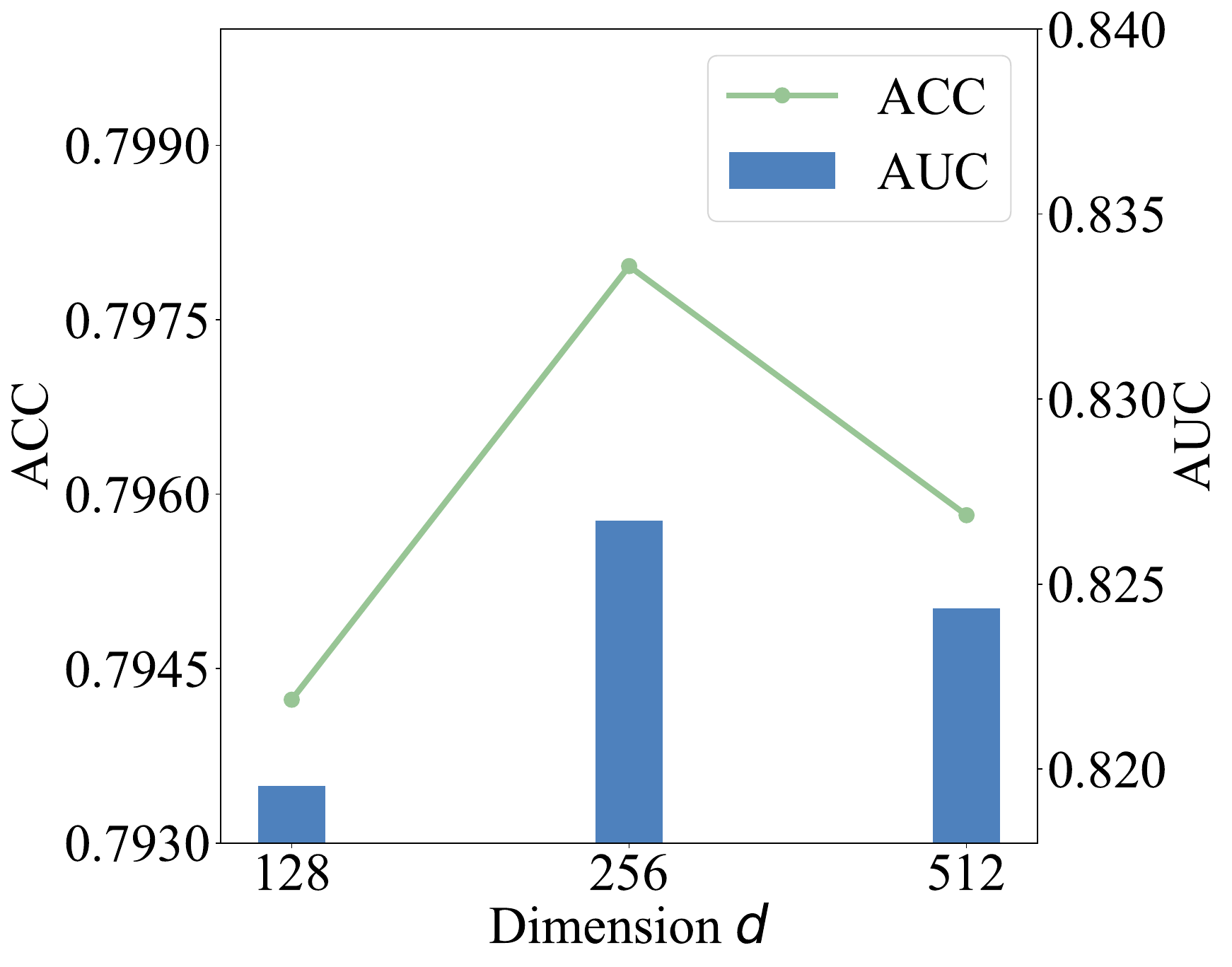}
        \caption{Programming.}
    \end{subfigure}
    \vspace{-5pt}
    \caption{Sensitivity analysis of representation dimension $d$ on four datasets.}
    \vspace{-8pt}
    \label{fig:sens_dim}
\end{figure*}

To investigate the performance of SINKT after introducing new questions, we remove 1/4 of the question from the training set and evaluate the model's prediction accuracy on these questions on the validation and test sets. Since the other three datasets do not provide textual information for the questions, we compare the performance of SINKT with that of DHKT and EERNNA (best and second best baselines) on the Programming dataset. We implement random embedding initialization for DHKT and EERNNA on those unseen questions to enable them to predict responses. The results of the experiment are shown in the Table\ref{tab:rq2}.
The experiment demonstrates that the SINKT performs better in predicting unseen questions, whereas the existing KT models struggle significantly with this task, achieving an AUC of less than 0.6.

We also investigate the cold start problem in real ITS.
We first organize an experiment of different training samples on ASSIST09. We choose $\{100, 500, 1000, 2000$, 2661 (all of the dataset)$\}$ students' learning history to train DHKT, EERNNA, and SINKT and evaluate them on the full test set. As shown in Figure~\ref{fig:sampleNum}, as the number of students decreases, performance decline of SINKT is more gradual compared to baselines, which indicates that SINKT can still learn accurate concept/question representations using textual and structural information even with limited training data.
Additionally, we present the training curves of DHKT, EERNNA, and SINKT on the ASSIST09 dataset in Figure~\ref{fig:trainCurve}. We argue that SINKT converges faster and exhibits minimal overfitting. This robustness in performance under limited data conditions underscores the effectiveness of incorporating rich textual and structural information in SINKT.

\begin{figure*}[t]
    \centering
    \begin{subfigure}{.296\linewidth}
    \centering
        \includegraphics[width=0.85\linewidth]{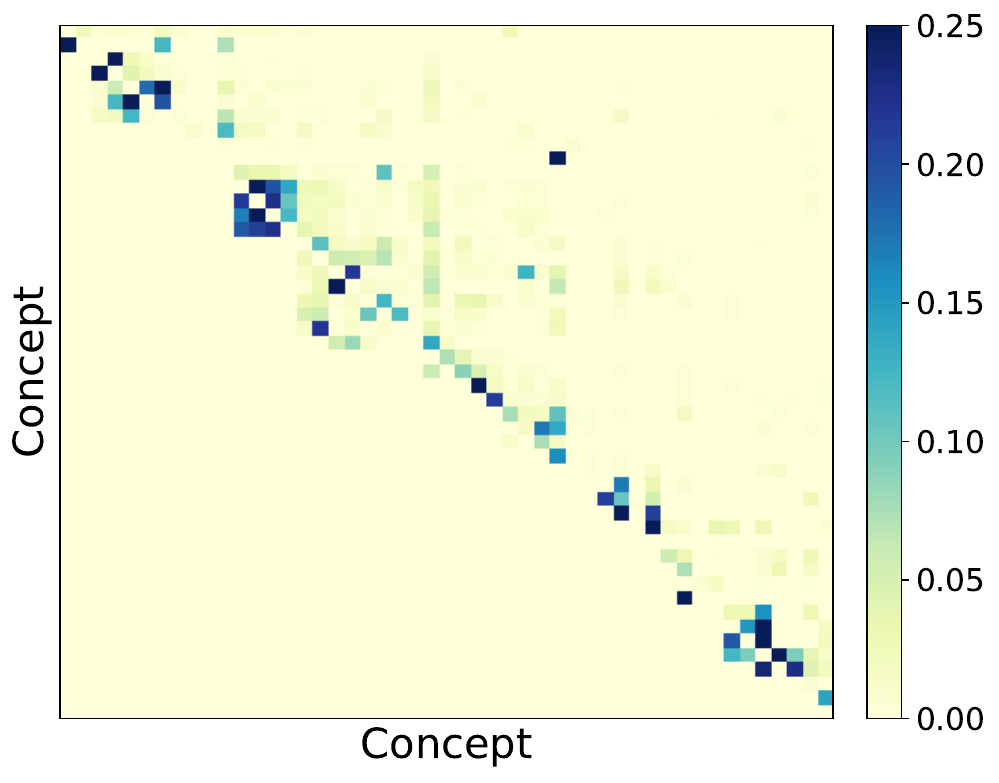}
        \caption{Transition graph.}
        \label{fig:transitionGraph}
    \end{subfigure}
    \begin{subfigure}{.29\linewidth}
    \centering
        \includegraphics[width=0.85\linewidth]{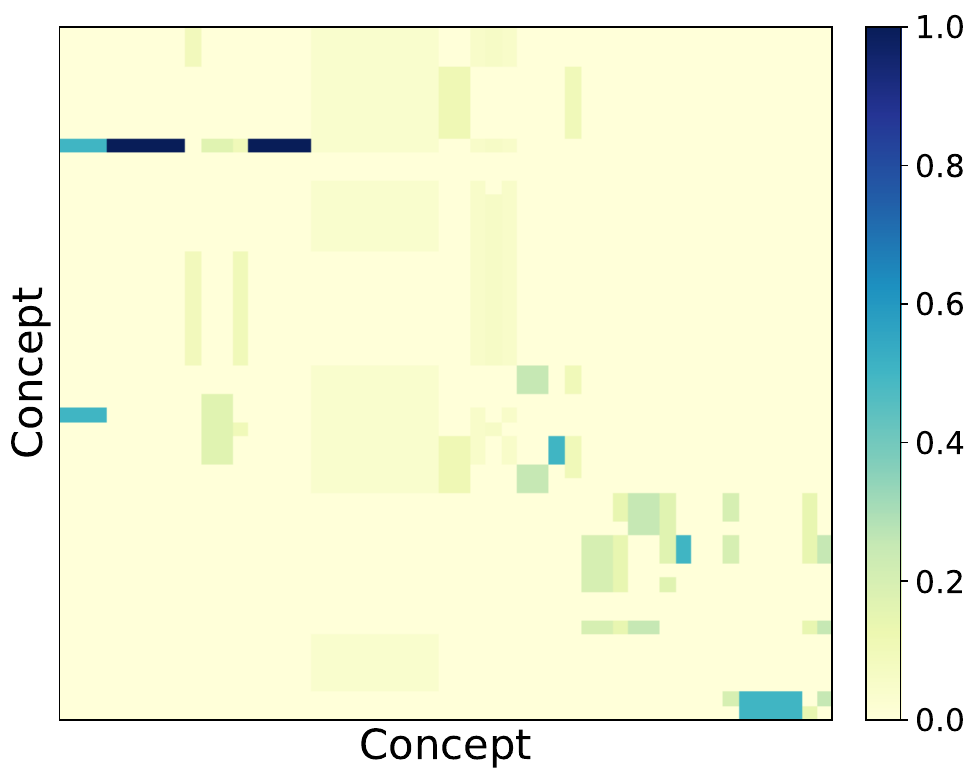}
        \caption{Concept graph generated by GPT-4.}
        \label{fig:genGraph}
    \end{subfigure}
    \begin{subfigure}{.296\linewidth}
    \centering
    \includegraphics[width=0.85\linewidth]{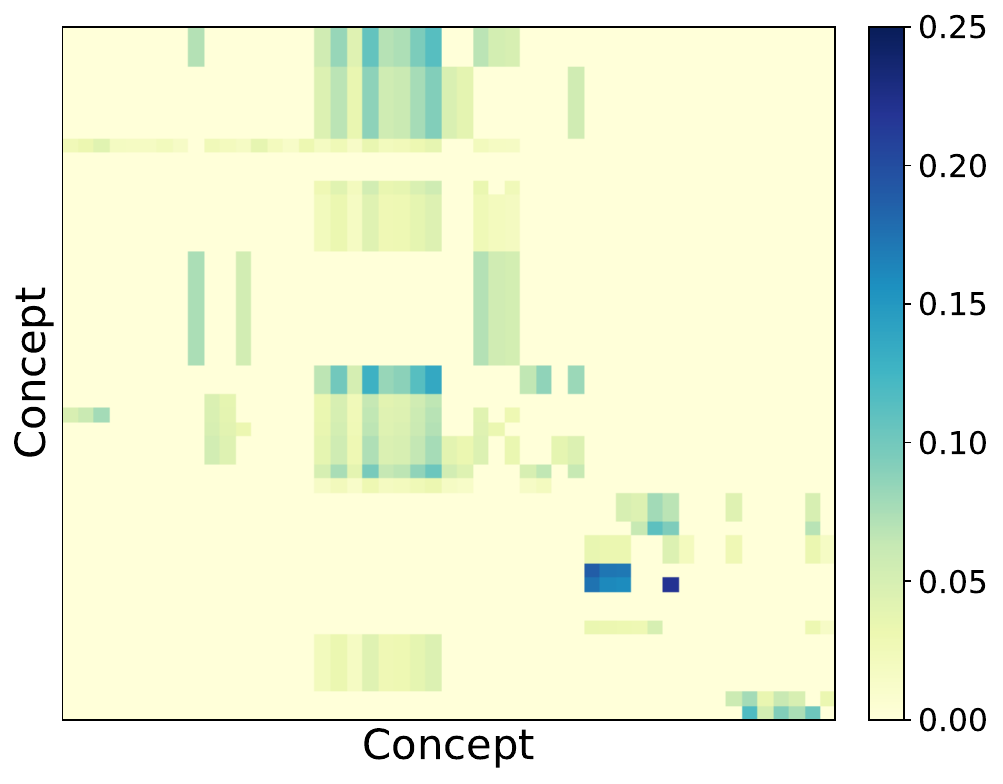}
        \caption{Concept graph with attention weights.}
        \label{fig:attentiongraph}
    \end{subfigure}
    \vspace{-5pt}
    \caption{Concept graph comparison between different generation types.}
    \vspace{-8pt}
    \label{fig:graphdemo}
\end{figure*}
\begin{figure}[t]
    \centering
    \includegraphics[width=0.85\linewidth]{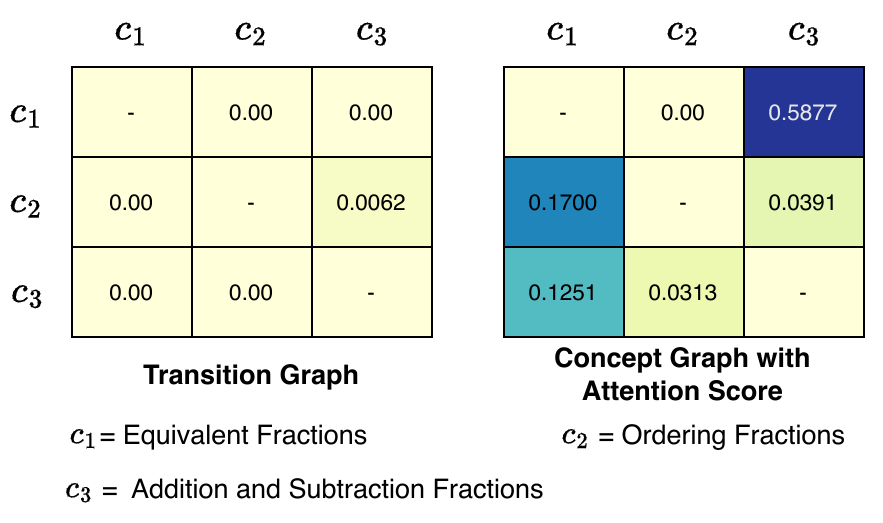}
    \vspace{-5pt}
    \caption{Detailed demonstration of transition graph and concept graph with attention score.}
    \vspace{-15pt}
    \label{fig:casestudy}

\end{figure}
\subsection{Ablation Study (RQ3)}
To further investigate the importance of each module in SINKT, we design three variants to conduct the ablation study, each of which removes or changes one part from the original SINKT:
\begin{itemize}[leftmargin=10pt]
    \item \textbf{SINKT-Linear} removes the jumping knowledge module in the graph encoder.
    \item \textbf{SINKT-GAT} removes GAT layers in the SIEnc, \textit{i.e. } only uses a linear layer to be a textual information adapter.
    \item \textbf{SINKT-Text} replaces the textual encoder with an embedding layer, which learns representations of questions and concepts from the dataset.
    \item \textbf{SINKT-Transition} replaces the concept generated by GPT-4 with the transition graph generated in the dataset.
\end{itemize}

Note that SINKT-Linear and SINKT-Graph still suitable for the inductive KT task, while SINKT-Transition and SINKT-Text do not support predicted responses to unseen questions.

We demonstrate the experimental results of SINKT and four variants in Table~\ref{tab:rq4}.
From the ablation study, we can derive several conclusions regarding the impact of different components in SINKT. Firstly, removing textual initialization (SINKT-Text) and the GAT layer (SINKT-GAT) in SIEnc results in a performance decline, indicating that both structural and textual information contributes to the learning of question and concept representations. Secondly, removing the jumping knowledge layer leads to a significant performance drop. This is because our heterogeneous graph is unevenly distributed, and the addition of the jumping knowledge layer allows the model to actively capture useful information. Thirdly, the performance degradation of SINKT-Transition demonstrates the effectiveness of the Concept Graph generated by GPT-4.

\subsection{Parameter Sensitivity Analysis (RQ4)}
In order to have an insight into the graph information conduction, we change the number of the Graph Encode layer $k$ ranging from 0 to 3.
Especially when $k=0$, the graph encoder degrades into a Linear layer without neighbor aggregation.
The experimental results on four datasets are shown in Figure \ref{fig:sens_layernum}.
We find that SINKT performs best in most cases when $k=2$, which indicates that second-order relationships in the concept-question heterogeneous graph are crucial for the learning process. The primary second-order relationships include concept-question-concept, concept-concept-question, and question-concept-question. In contrast, lower-order relationships provide limited neighbor information, while higher-order relationships introduce noise.

Furthermore, we investigate the sensitivity of dimension $d$ of representations vectors $\Tilde{q},\Tilde{c}$. We evaluate SINKT's performance on three different numbers of $d$: $\{128,256,512\}$.
The experimental results on four datasets are shown in Figure~\ref{fig:sens_dim}. We suggest that the best representation dimension of SINKT is 256. However, existing transductive KT models usually choose $64$ or $128$ as the representation dimension of concepts and questions. The gap in the selection of dimension values primarily arises from the introduction of open-world knowledge in SINKT. This knowledge is extensive and requires additional dimensions to be effectively encoded.

\subsection{Quality of the Graph Generation (RQ5)}

As we generate a concept relation graph by GPT-4, it is critical to inspect the quality of the graph.
To investigate the overall quality of the concept-concept graph, we compare the transition graph and our concept relation.
Transition graph denotes the transition probability of concept answering, which is derived from dataset and could not be generated without training data.
We demonstrates the transition graph of ASSIST09 in Figure~\ref{fig:transitionGraph}. The weight in the $i$-th row and the $j$-th column of the transition graph means the probability of concept $c_i$ appears after concept $c_j$ in the dataset. 
Since the learning sequence of concepts are generated by ITS's recommendation algorithm, the distribution of prerequisite concepts in  the transition graph is relatively fixed.
The concept relation graph generated by GPT-4 is shown in Figure~\ref{fig:genGraph}.
The weight of $c_i$ in the figure is $\frac{1}{\mid \mathcal{N}_{c_i}^c\mid}$.
Meanwhile, we visualize the attention score of the first GAT layer in Figure~\ref{fig:attentiongraph}.
Compared to transition graph, our concept graph with attention weight is more integrated.

We also give a case study for the concept graph's rationality in Figure~\ref{fig:casestudy}.
In the transition graph, concept $e_1,e_2,e_3$ nearly do not have correlations. While in our concept graph, relationship between every two concepts are clear and can be judged by reasonable scores.

\section{Conclusion}
In this paper, we introduce a structure-aware knowledge tracing framework SINKT with large language model, which are suitable for both transductive and inductive KT tasks. SINKT leverages LLMs to build structural relationships between concepts and questions and encode semantic information. To better aggregate these information, in the learning process of students, we carefully design a textual information encoder, a structural information encoder and a student state encoder.
Experimental results on four real-world datasets show that SINKT outperforms current ID-based KT models on both transductive and inductive KT tasks.
We also provide case studies and ablation studies to prove the effectiveness of each module in SINKT.
In the future, we will try to incorporate more wealthy open-world knowledge and make more accurate predictions on inductive KT tasks.

\begin{acks}
    The Shanghai Jiao Tong University team is partially supported by National Natural Science Foundation of China (62177033, 62076161) and Shanghai Municipal Science and Technology Major Project (2021SHZDZX0102). The work is also sponsored by Huawei Innovation Research Program.
 We thank MindSpore~\cite{mindspore} for the partial support of this work.
\end{acks}
\newpage


\end{document}